\documentclass[final,onefignum,onetabnum]{siamonline250106}

\ifpdf
\hypersetup{
  pdftitle={Multiplicative Reweighting for Robust Neural Network Optimization},
}
\fi

\usepackage{booktabs}
\usepackage{amsmath}
\usepackage{amssymb}
\usepackage{mathtools}

\usepackage{multirow}
\usepackage{adjustbox}
\usepackage{float}
\usepackage{color}

\usepackage{physics}

\usepackage{url}            % simple URL typesetting
\usepackage{amsfonts}       % blackboard math symbols
\usepackage{nicefrac}       % compact symbols for 1/2, etc.
\usepackage{microtype}      % microtypography

\usepackage{wrapfig}
\RequirePackage[format=plain,labelformat=simple,labelsep=period,font=small,compatibility=false]{caption}
\RequirePackage[font=footnotesize,skip=3pt,subrefformat=parens]{subcaption}

\usepackage{algorithm}
\usepackage{algorithmic}

\DeclareMathOperator{\E}{\mathbb{E}}

\DeclareMathOperator*{\argmin}{arg\,min}
\newcommand{\yT}{\Tilde{y}}
\newcommand{\LT}{\Tilde{L}}
\newcommand{\lT}{\Tilde{\ell}}

\newcommand{\bs}[1]{\boldsymbol{\rm {#1}}}

\title{Multiplicative Reweighting for Robust Neural Network Optimization
}

\author{Noga Bar\thanks{Tel Aviv University (\email{nogabar@mail.tau.ac.il}).}
\and Tomer Koren \thanks{ Tel Aviv University and Google.}
\and Raja Giryes\thanks{Tel Aviv University }}

\headers{Multiplicative Reweighting for Robust Neural Network Optimization}{}

\ifpdf
\hypersetup{ pdftitle={Multiplicative Reweighting for Robust Optimization} }
\fi

\begin{document}

\maketitle

\begin{abstract}
Neural networks are widespread due to their powerful performance.
Yet, they degrade in the presence of noisy labels at training time.
Inspired by the setting of learning with expert advice, where multiplicative weights (MW) updates were recently shown to be robust to moderate data corruptions in expert advice, we propose to use MW for reweighting examples during neural networks optimization.
We theoretically establish the convergence of our method when used with gradient descent and  prove its advantages in 1d cases.
We then validate empirically our findings for the general case by showing that MW improves neural networks' accuracy in the presence of label noise on CIFAR-10, CIFAR-100 and Clothing1M. We also show the impact of our approach on adversarial robustness.
\end{abstract}



\section{Introduction}
\label{sec:introduction}

Deep neural networks (DNNs) have gained massive popularity due to their success in a variety of applications.
Large accurately labeled data are required to train a DNN to achieve good prediction performance.
Yet, such data is costly and require a significant amount of human attention. To reduce costs, one may train a network using annotated datasets that were created with lesser efforts, but these may contain label noise which impedes the training process. 

We study methods to mitigate the harmful effect of noise in the training process.
Motivated by learning with expert advice, we employ Multiplicative-Weight (MW) updates ~\cite{littlestone1989weighted,freund1997decision} for promoting robustness in the training process.
In online learning, it was shown theoretically that MW achieves optimal regret in a variety of scenarios, e.g., with losses drawn stochastically~\cite{mourtada2019optimality} or adversarially~\cite{littlestone1989weighted,freund1997decision}, and also in 
an intermediate setting with stochastic losses that are moderately corrupted by an adaptive adversary~\cite{amir2020prediction}.
Thus, it is natural to consider this technique in a new context of DNN training with noise, such as label noise in train data.

\begin{figure}[t]
    \centering
    \includegraphics[width=0.5\linewidth]{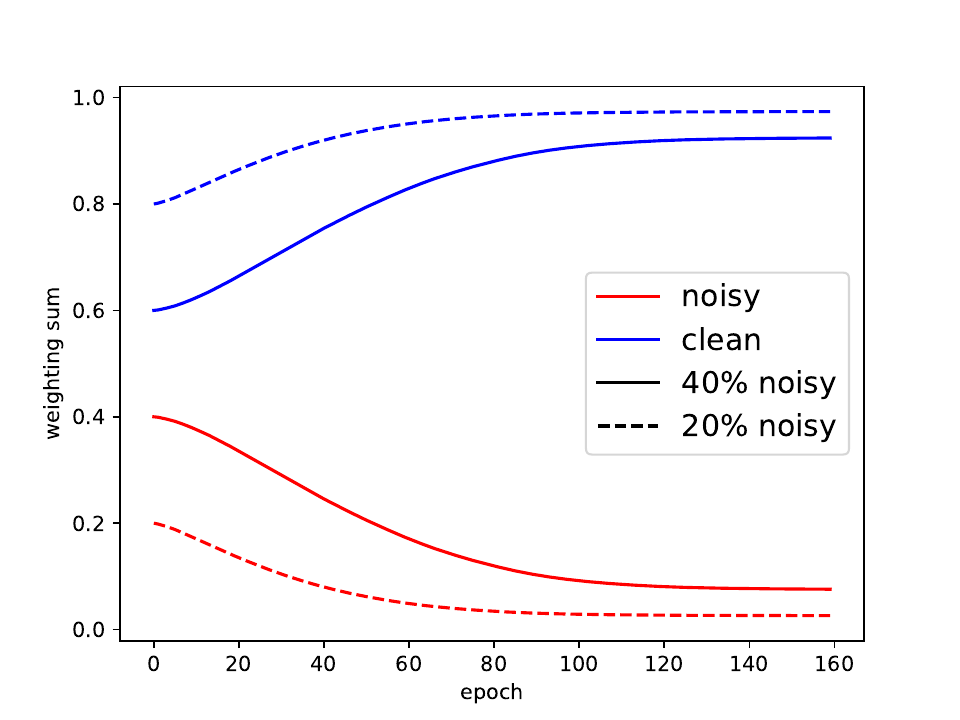}
    \caption{Evolution of the multiplicative weights sum for clean and noisy examples with 20\% and 40\% label noise. The sum of all weights is 1. Note how the noisy examples weight decreases through the training and thus they less affect the network.}
    \label{fig:teaser}
\end{figure}
 
To employ MW for this purpose, we interpret the examples as experts and the predictions induced by the network as their advice.
We then encounter a loss function over the predictions, which facilitates the use of MW.
Thus, instead of learning by minimizing a uniform average of the losses, we propose to learn a weighted version of the latter in which not all examples affect the training process uniformly.
We employ a MW update rule for reweighting the examples during training, which we denote as multiplicative reweighting (MR). It alternates between SGD steps for optimizing the DNN's parameters and MW updates for reweighting.
\cref{fig:teaser} shows the weighting evolution during training where noisy examples have notably lower weights than their ratio in the training data.
MR is simple, generic and can fit easily in most training procedures.
We establish the convergence of a simplified version of MR under mild conditions and prove the efficacy of MR for 1d input data.

Motivated by our theoretical findings, we show empirically that MR is also beneficial in the realistic high-dimensional case for training with noisy labels.
We demonstrate its advantage using two popular DNN optimizers: SGD (with momentum) and Adam.
We evaluate our approach both on artificial label noise (symmetric, pair-flip and instance dependent) in CIFAR-10 and CIFAR-100, and real noise in Clothing1M~\cite{xiao2015learning}.
We compare to common techniques: Mixup \cite{zhang2017mixup} and label-smoothing \cite{szegedy2016rethinking}, which are known to improve accuracy with label noise, and the state-of-the-art sparse regularization \cite{zhou2021learning,karim2022unicon}. We improve performance when we combine MR with them. We further compare to \cite{arazo2019unsupervised,yao2020searching,han2018co,Liu16Classification,shu2019meta}, which also were suggested to handle label noise.

In the setting of learning with expert advice MW is known to be optimal in the presence of worst-case adversarial loss. 
When used with neural networks, we show that MR leads to a better effective Lipschitz constant in the 1d case than GD.
Note that a low Lipschitz constant, has been shown to improve robustness to adversarial noise \cite{jakubovitz2018improving,Ross2017InputGradients}.
 This suggests that MR may be useful for improving adversarial robustness in neural networks.
Hence, we also test MR in the context of adversarial attacks. We show how MR can improve adversarial training, which is known to be one of the best approaches against adversarial attacks. Specifically, we show the MR advantage with Free Adversarial Training \cite{shafahi2019adversarial}, TRADES \cite{zhang2019theoretically}, LAS \cite{jia2022adversarial} and MAIL \cite{liu2021probabilistic}.

Overall, we show that our method can fit into the deep learning optimization toolbox as an easy to use tool that can help improving network robustness in a variety of scenarios and training pipelines. Our code appears in \url{https://github.com/NogaBar/mr_robust_optim}.

\section{Related Work}

{\bf Multiplicative weights.} 
We are inspired by recent research on corruption-robust online learning, and specifically by new analyses of the multiplicative weight (MW) algorithm in the setting of learning with expert advice.
Online learning is concerned with minimizing the regret which is the difference between cumulative loss of the learner's actions and that of the best action in hindsight.
In this context, MW~\cite{littlestone1989weighted,freund1997decision} is known to achieve optimal regret (of the form $\Theta(\sqrt{T \log N})$, where $T$ is the time horizon and $N$ is the number of experts), in a general setup where the losses are chosen by an adversary.
Very recently, MW was shown to achieve a constant regret (i.e., independent of the time horizon) when the losses are i.i.d.~\cite{mourtada2019optimality}, and that this bound deteriorates very moderately when some of the stochastic losses are corrupted by an adversary~\cite{amir2020prediction}.
Analogous results have been established in related settings, e.g., in Multi-armed Bandits~\cite{lykouris2018stochastic,gupta2019better,zimmert2019optimal}, Markov Decision Processes~\cite{lykouris2019corruption}, and more.
Note that the regret bound in the online learning setting with stochastic losses can be used to bound the associated expected optimization loss, which is the object of interest in supervised and stochastic learning settings.
Ideas from online learning and regret minimization were used in other contexts in deep learning, e.g., for improving neural architecture search~\cite{nayman2019xnas}, online deep learning~\cite{sahoo2017online}, and robust loss functions design~\cite{amid2019robust}. Recently, a multiplicative update rule of the network parameters was used to induce compressible networks \cite{bernstein2020learning}. 

{\bf Noisy Labels.}
Many methods were proposed for learning with noisy labels \cite{song2020learning}. 
It was suggested to adjust the network architecture \cite{xiao2015learning,Goldberger17Training,han2018co,wei2020combating,karim2022unicon}, learn with a suitable loss function \cite{Ghosh15Making,Ghosh17Robust,Malach17Decoupling,Li17Learning,han2018masking,Tanaka18Joint,zhang2018generalized,xu2019l_dmi,Thulasidasan2019CombatingLN,Han2019Deep,Shen19Learning,amid2019two,amid2019robust,wang2019symmetric,zhou2021learning,liu2022robust} and employ semi-supervised techniques \cite{Vahdat17Toward,zhang2020distilling,nguyen2020self,li2020dividemix,liu2020early,li2022selective}.
Some works tried to analyze factors that impact the network robustness to noise \cite{Krause16unreasonable,SunSSG17,Ma18Dimensionality,li2020gradient,drory2020resistance} and other suggested choosing a partial subset of the parameters in the network to update \cite{bai2021understanding,xia2021robust}.

There are techniques that were not necessarily designed to training with noisy labels but significantly improve over the vanilla optimization, for example, SAM \cite{foret2020sharpness}, mixup \cite{zhang2017mixup} and label smoothing \cite{szegedy2016rethinking,lukasik2020does}.

The connection between the label noise problem and robustness against adversarial attacks was previously studied.
On the one hand, it was shown that label noise harms network robustness. However, removing the noise is not sufficient to produce robustness \cite{sanyal2020benign}.
On the other hand, adversarially robust training reduces vulnerability to noisy labels. Furthermore, examples that are easy to attack are more likely to suffer from label noise \cite{zhu2021understanding}

{\bf Loss Weighting in DNNs.} We elaborate on methods that use weighting to handle noisy labels (among other problems).
There are two main weighting approaches. One is loss weighting, where different classes affect the loss non-uniformly.
This strategy was studied theoretically \cite{natarajan2013learning} and empirically \cite{lin2017focal, patrini2017making, hendrycks2018using,xia2019anchor}.
Note that the weighting of each class may be learned in an instance-dependant manner
\cite{xia2020part,yao2020dual,yang2021estimating,li2022estimating}.

The second approach is weighting examples, either locally within a mini-batch \cite{jenni2018deep,rozsa2018towards} or globally, weighting all examples in the train set.
Global weighting is usually done during the optimization of the DNN.
These techniques include learning to weight with gradient methods \cite{ren2018learning,shu2019meta,zhang2021learning}, exploiting the variance and of the DNN's features and predictions \cite{chang2017active,Yao19Deep}, employ graph NN and analyse the structural relations of inputs \cite{zhang2021dualgraph}, using sample neighbourhood \cite{Guo18CurriculumNet} or importance sampling for re-weighting the examples \cite{Liu16Classification,wang2017multiclass}, weighting the examples according to beta mixture model \cite{arazo2019unsupervised} and  exploit conditional value-at risk \cite{soma2020statistical}. 
Some of those works are used for evaluating label noise in the data \cite{Liu16Classification,arazo2019unsupervised} and in others it was shown that the methods convergence to a critical point \cite{ren2018learning,shu2019meta,soma2020statistical}. 

 There are previous works that suggested using weighting as a measure of confidence for each example and learning a more robust network \cite{liu2021probabilistic,zhang2020geometry}.

Our approach differs from existing works in that we use global weighting during optimization based on MW. We also theoretically establish the convergence of our methods and demonstrate in 1d cases the possible benefits of MW. Note that some of the above techniques, unlike our method, require an additional clean set, for performing the weighting.

\textbf{Curriculum Learning. } Another concept of ordering (or weighting) the importance of examples is curriculum learning \cite{bengio2009curriculum} or self-paced learning \cite{kumar2010self,jiang2015self} which aims to improve the learning process by ordering carefully the input examples. Initially, the order was established using prior expert data where tasks that are perceived by humans as easy are given to the learning model first \cite{bengio2009curriculum}. Later it was suggested to employ learning techniques to perform the ordering \cite{fan2018learning}. 
This concept is widely used to tackle label noise \cite{jiang2018mentornet,yao2020searching,lyu2019curriculum,zhou2020robust,xia2021sample} and there are various methods for ordering. Sometimes, curriculum learning is combined with other techniques such as semi-supervised learning \cite{jiang2018mentornet,zhou2020robust}.
Our approach can be viewed as a curriculum learning method, where MW is used to determine in an unsupervised manner the importance of each example.

\begin{algorithm}[t]
   \caption{Multiplicative Reweighting (MR) with SGD}\label{alg:sgd_mw_epoch}
\begin{algorithmic}
   \STATE {\bfseries Input:} data $\bs{x_i}$, $y_i$ for $1 \leq i \leq N$, $\alpha$- GD step size and $\eta$- MW step size, $B$-batch size, $\mu$ - maximal weight constrain.
   \STATE  \textbf{Initialize:} $w_{0,i}=1$, $\theta_0^{N/B}$ network initialization.
   \FOR{$t=1$ {\bfseries to} $T$}
%   \STATE \textbf{Normalize P:} $p_{t,i} = \frac{w_{t,i}}{\sum_{j=1}^N w_{t,j}}$
   \STATE \textbf{Project:} $\displaystyle p_{t} =$ Project$\displaystyle (w_{t}, \mu)$, \cref{alg:projection}. \COMMENT{{\color{black}Project $w_t$ onto valid probabilities.}}
   \STATE \textbf{Batched SGD epoch:} 
   \FOR{$m=1$ \bfseries{to} $N/B$}
   \STATE $\displaystyle \theta_t^0 = \theta_{t-1}^{N/B}$
   \STATE $\displaystyle \theta_t^m = \theta_t^{m-1} -\alpha \frac{\sum_{i\in B_m^t} p_{t,i} \nabla \ell_i(\theta_t^{m-1})}{\sum_{i\in B_m^t} p_{t,i}}$
   \ENDFOR
   
   \STATE \textbf{MW}:~$\displaystyle w_{t+1,i} = \exp(-\eta \sum_{s=1}^{t} \ell_i(\theta_s^{N/B}))$ \COMMENT{{\color{black}Update weights via cumulative losses.}}
   
   \ENDFOR
\end{algorithmic}
\end{algorithm}

\section{Training with Multiplicative Reweighting}
In the common neural network training setup for classification, we have $N$ training examples $\bs{x_i} \in \mathbb{R}^d$ with labels  $y_i\in \mathbb{R}$ ($1 \leq i \leq N$) that are sampled from some unknown distribution $\mathbb{P}[\bs{X}, Y]$. Denote the loss function for example $i$ as $\ell_i(\theta) = \ell(\theta;\bs{x}_i,y_i)$ where $\theta$ is the network's parameters.

Most DNNs are trained with gradient methods and assign the same importance to all samples in the train set. They usually use Empirical Risk Minimization (ERM) \cite{shalev2014understanding} that treats all examples equally:
\begin{align*}
    \theta^\star \in \argmin_{\theta} \frac{1}{N} \sum_{i=1}^N \ell_i(\theta).
\end{align*}
Yet, we have a reason to suspect that not all examples are reliable with the same confidence so we want to learn a weighted (non-uniformly averaged) loss function:
\begin{align}\label{eq:min_weight_loss}
     p^\star, \theta^\star \in \argmin_{\genfrac{}{}{0pt}{}{\theta, p\in \Delta_N}{p_i < \mu/N},}  \sum_{i=1}^N p_i \ell_i(\theta),
\end{align}
% \vskip -0.1cm
where $\Delta_N$ is the $N$ dimensional simplex and $p^\star$ is supposed to match the reliability of the examples.
Thus, finding $p^\star$ requires learning the examples that are likely in $\mathbb{P}[\bs{X}, Y]$ without any extra data.

We can observe the problem of finding $p^\star$ from a different perspective: the examples $\bs{x_i}$ can be seen as experts, the predictions they produce by a forward pass as their advice and thus, by applying a loss function over the advice we can solve the problem under the setting of learning with expert advice.
In the online learning setting, MW is a simple and well established algorithm known to produce optimal results in the context of expert advice when the true losses are stochastic and adversarial. This motivates us to use MW also with neural networks using our proposed perspective.
The MW algorithm assigns a weight to each expert according to their cumulative loss, in a way that weights are negatively proportional to the losses. Hence, the learner gradually overlooks experts that perform poorly.

Another motivation for using MW for improving network robustness is that in the presence of label noise, we know that DNNs optimized with SGD and cross entropy loss tend to learn first from clean examples and at the end of the training process they overfit the corrupted examples \cite{arazo2019unsupervised,zhang2018generalized,liu2020early}.
The corrupted examples are not typical to their class and suffer high loss values throughout a long period of training.
The MW rule decays exponentially the weights of examples with high loss and treats them as bad experts. Thus, we expect an improved accuracy. Our method is not expected to harm the performance on the clean data since in the general case where there is no known label noise, one can train a network without hard or forgettable examples and still achieve a good accuracy \cite{swayamdipta2020dataset,toneva2018empirical}.

MW outputs a distribution $p$ over the experts and requires loss per expert as feedback.
Thus, we suggest to use the losses $\ell_i(\theta)$ produced by the network's parameters $\theta$.
Hence, a change in $\theta$ yields different losses which leads to new MW update.
In order to learn both $\theta$ and $p$ as in \cref{eq:min_weight_loss} we propose to alternate between SGD for optimizing the network parameters $\theta$ and  MW for calculating the distribution $p$. 
We denote this approach multiplicative reweighting (MR) and it is summarized in \cref{alg:sgd_mw_epoch}.

To update $\theta$, we use a full epoch of batched gradient updates with mini-batch size $B$ using the weighted loss (as in \cref{eq:min_weight_loss}) according to $p_t$, the distribution at epoch $t$.
Using the a weighted loss version leads to variations in the magnitude of updates over different mini-batches.
To keep the learning rate stable across mini-batches, we normalize the update by the sum of weights $\sum_{i\in B_m^t} p_{t,i}$ in the mini-batch $B_m^t$, where $m$ is the mini-batch index at epoch $t$. 

For updating the probability $p_{t}$ of the examples at epoch $t$, we use the weights $w_{t}$ that are calculated using the MW update (based on the DNN parameters in the previous epoch) and project them to the constrained domain to make $p_t$ a valid distribution.
The weights for $t+1$ are updated according to all available loss history including $\ell(\theta_{t}^{N/B})$. In the MW step, we have the step-size $\eta$ that controls the non-uniformity of the distribution.
Note that the MW update requires $\ell_i(\theta)$ for $1\leq i \leq N$ so an additional forward pass is performed at the end of each epoch.
 While it is true that simple methods for handling label noise—such as Mixup, random weighting, and label smoothing—introduce negligible additional computational overhead, other approaches like Co-Teaching \cite{han2018co} and Meta-Weight \cite{shu2019meta} require training two models in parallel, and Arazo et al. \cite{arazo2019unsupervised} involves additional forward passes. An empirical evaluation of training time is provided in \cref{tab:time}.

For avoiding learning degenerated probability such that few examples have high probability, we limit the highest possible probability by adding to \cref{eq:min_weight_loss} a linear constrain of the form $p_i < \frac{\mu}{N}$ for all $i$, and for a given parameter $\mu$.
 
To maintain the linear constraint, we use a projection algorithm which projects the probabilities that were derived from the recent MW update step into a limited domain in the simplex with non-degenerated probabilities.
In the supplementary material attached to this paper, we present \cref{alg:projection}, where we detail the procedure of projecting the raw weighting $w_i$ to the constrained simplex with respect to the KL-divergence \cite{warmuth2008randomized,koolen2010hedging}.
First, the weights are normalized to be in the simplex, then all the excess mass of weights with more than $\mu/N$ weight is redistributed according to the magnitude of the rest of the weights.
In this way we redistribute the excess mass in a way that changes the least the value of the original KL-divergence of the distribution achieved by MW while maintaining the maximal weight constrain.
The process is performed iteratively until all weights are below the chosen threshold and its total complexity is $O(N^2)$.

In \cref{sec:exp}, we explore the need of the projection and find that the weights are in the desired range without the projection and thus in practice it is not required.
Practically, the network does not assign significantly lower losses to a small set of examples during a long period of the training.
We suspect the reason for it is the training's stochasticity which induces high loss variability across epochs.

\begin{algorithm}[t]
  \caption{Multiplicative Reweighting (MR) with Gradient Descent}\label{alg:gd_mw}
\begin{algorithmic}
   \STATE {\bfseries Input:} data: $\bs{x_i}$, $y_i$, size $N$, $\alpha$- GD step size and $\eta$- MW step size.
   \STATE {\bf Initialize:} $w_{0,i}=1$ and $\theta_0$ - the network initialization, $\mu$ - maximal weight constrain.
   \FOR{$t=1$ {\bfseries to} $T$}
   \STATE \textbf{Project:} $\displaystyle p_{t} =$ Project$\displaystyle (w_{t}, \mu)$, \cref{alg:projection}. \COMMENT{{\color{black}Project $w_t$ onto valid probabilities.}}
   \STATE \textbf{GD:} $\theta_{t+1} = \theta_t - \alpha \sum_{i=1}^N p_{t,i}  \nabla \ell_i(\theta_t))$
   \STATE \textbf{MW:} $w_{t+1,i} = \exp(-\eta \sum_{s=1}^{t+1} \ell_{i}(\theta_s))$ \COMMENT{{\color{black}Update weights via cumulative losses.}}
   \ENDFOR
\end{algorithmic}
\end{algorithm}

\begin{algorithm}[t]
   \caption{MR with LS loss (for the 1d example)}
   \label{alg:ls_mw}
\begin{algorithmic}
   \STATE {\bfseries Input:} data: $\bs{x}$, $\bs{\yT}$, and $\eta$- MW step size.
   \STATE \textbf{Initialize:} $w_{0,i}=1$, $\theta_0 = 0$.
   \FOR{$t=1$ {\bfseries to} $T$}
   \STATE \textbf{Normalize P:} $p_{t,i} = \frac{w_{t,i}}{\sum_{j=1}^N w_{t,j}}$, update probability matrix: $\bs{P_t} = \text{diag}(p_t)$ 
   \STATE \textbf{LS:} $\theta_{t+1} = \bs{x \sqrt{P_t} (\sqrt{P_t}x^T x \sqrt{P_t})^\dagger \sqrt{P_t}\yT }$
   \STATE \textbf{Calculate Loss}: $\lT_{t+1,i} = \frac{1}{2}(\theta_{t+1} x_i - \yT_i)^2$
   \STATE \textbf{MW:} $w_{t+1,i} = \exp(-\eta \sum_{s=1}^{t+1} \lT_{s,i})$ 
   \ENDFOR
\end{algorithmic}
\end{algorithm}

\section{Theoretical Analysis}
We turn to provide theoretical guarantees for MR. 
For the simplicity of the analysis, we focus mainly on its version with full batch GD but demonstrate how to extend to SGD as well.
We start with MR with GD as described in \cref{alg:gd_mw}. 
We prove convergence to a critical point both for \cref{alg:gd_mw} and a simplified stochastic variant of it.
To prove the advantage of our method with noisy labels, we analyse two 1d illustrative examples.
The first is a logistic regression learned with a full batch as in \cref{alg:gd_mw}.
The second is linear regression, where we use MR with a least squares loss and show convergence to the optimal solution.
All proofs and definitions of $\beta$-smoothness, $G$-lipschitz and $B$-bounded function appear in the supp. mat.

\subsection{Convergence}
It is well-known that when GD is applied on a $\beta$-smooth function, it converges to a critical point (e.g., \cite{bubeck2015convex}).
We thus start with an analysis of the full-batch version of our method (\cref{alg:gd_mw}) and show that it also converges to a critical point with the same assumptions. 

To establish convergence, we first show an equivalent of the descent lemma proving that in each step the weighted loss is decreasing.

\begin{lemma}[Equivalence to Descent Lemma] \label{lemma:descent_lemma} For a $\beta$-smooth loss $\ell(\cdot)$, and $\theta_{t+1}$, $p_{t+1}$ updated as in \cref{alg:gd_mw} with GD step size of $\alpha=\frac{1}{\beta}$ and MW step size $\eta > 0$
\begin{align*}
    \sum_{i=1}^N p_{t+1, i} \ell_i(\theta_{t+1}) -  p_{t, i}  \ell_i(\theta_{t})
    \leq -\frac{1}{2\beta} \norm{\sum_{i=1}^N p_{t,i} \nabla \ell_i(\theta_t)}^2&.
\end{align*}
\label{descent_lemma}
\end{lemma}

The lemma establishes that the loss of each two consecutive iterations in  \cref{alg:gd_mw} is non-increasing.
Now, we can state formally and prove convergence to a critical point with GD.

\begin{theorem}[convergence]
\label{thm:gd_convergence}
For a $\beta$-smooth loss $\ell(\cdot)$, and $\theta_{t+1}$, $p_{t+1}$  being updated according to MR with GD (as in \cref{alg:gd_mw}) with step size $\alpha=\frac{1}{\beta}$ and MR step size $\eta > 0$, we have that
\begin{align*}
    \frac{1}{T} \sum_{t=1}^T & \norm{\sum_{i=1}^N p_{t,i} \nabla \ell_i(\theta_t)}^2 \leq 
    \frac{2\beta}{T} \bigg(\frac{1}{N}\sum_{i=1}^N \ell_i(\theta_0) - \sum_{i=1}^N p_i^\star \ell_i(\theta^\star)\bigg),
\end{align*}
where $p^\star, \theta^\star \in \argmin_{\genfrac{}{}{0pt}{}{p \in \Delta_N ,\theta}{p_i < \mu/N}} \sum_{i=1}^N p_i \ell_i(\theta)$.
\end{theorem}
The theorem states that the average over the time horizon $T$ of the gradients of the weighted loss goes to $0$. The convergence rate depends on constants, which are determined by the smoothness $\beta$, the starting point $\theta_0$ and the optimal parameters $p^\star$, $\theta^\star$.

\cref{alg:gd_mw} may converge to a degenerate solution, where only one weight is non-zero. Thus, in the supp. mat., we also prove convergence when the distribution $p$ is constrained and each example has a minimal weight (in \cref{alg:gd_mw} the minimal weight is $0$).

As the average of gradients is bounded as \cref{thm:gd_convergence} states, we can run \cref{alg:gd_mw} and achieve at large enough time $t\in[T]$ any small size of gradient we want.
It essentially means that  $\exists t\in[T]$ for which the parameters $\theta_t$ and $p_t$ are as close as we want to a critical point if such exists.
\begin{corollary}[{\color{black}convergence rate}]
Under the same conditions of \cref{thm:gd_convergence}, for $\tilde{T} = O\big(\frac{\beta }{\epsilon}\big(\sum_{i=1}^N p_{0,i} \ell_i(\theta_0) - \sum_{i=1}^N p_i^\star \ell_i(\theta^\star)\big)\big)$ iterations of \cref{alg:gd_mw}: $\norm{\sum_{i=1}^N p_{t,i} \nabla \ell_i(\theta_t)}^2 \leq \epsilon$ for $t \ge \tilde{T}$.
\end{corollary}

We turn to analyse a stochastic version of the training, where instead of updating $\theta_{t+1}$ with respect to all examples, we sample one example $i_t$ each time according to the current distribution over the examples $p_t$.
We update $\theta_{t+1}$ based on the $i_t$th example's gradient, i.e, $\theta_{t+1} = \theta_t - \alpha \nabla \ell_{i_t}(\theta_t)$.
Namely, we analyse a variant of \cref{alg:gd_mw} that replaces GD with a SGD variant that samples according to $p_t$ instead of sampling the examples uniformly.

\begin{theorem} [SGD convergence]\label{thm:sgd_mw_converge}
For $1\leq i \leq N: \ell_i(\cdot)$ is $\beta$-smooth, $G$-lipschitz $B$-bounded. Define $F(\theta, p) = \E_{i\sim p} \ell_i(\theta)$.
Then when running MR with SGD that samples according to $i_t \sim p_t$ and uses $\alpha=\sqrt{\frac{2B}{G^2\beta T}}$, it holds that: $\frac{1}{T} \sum_{t=1}^T \E_{i_1, ..., i_t} \norm{\nabla F(\theta_t, p_t)}^2 \leq G \sqrt{\frac{2\beta B}{T}}.$
\end{theorem}

Here we further use the common assumptions that the loss is bounded for simplicity of presentation and also that it has bounded gradients.
The above theorem bounds the average of the expectations of the gradients (instead of the actual gradients as in \cref{thm:gd_convergence}).
The key idea in the proof is that the expectation of the gradient at time $t$ is an unbiased estimator to the average gradient $\sum_{i=1}^N p_{t,i} \nabla \ell_i(\theta_t)$.

Our analysis focuses on convergence of worse-case and we achieve a similar rate to the common algorithms GD and SGD under the same assumptions of the loss function~\cite{shalev2014understanding}.

\subsection{Guarantees for 1d Case with Label Noise} \label{sec:illustrative}
\begin{figure*}
    \centering
    \begin{subfigure}{0.32\linewidth}
    \includegraphics[width=1\linewidth]{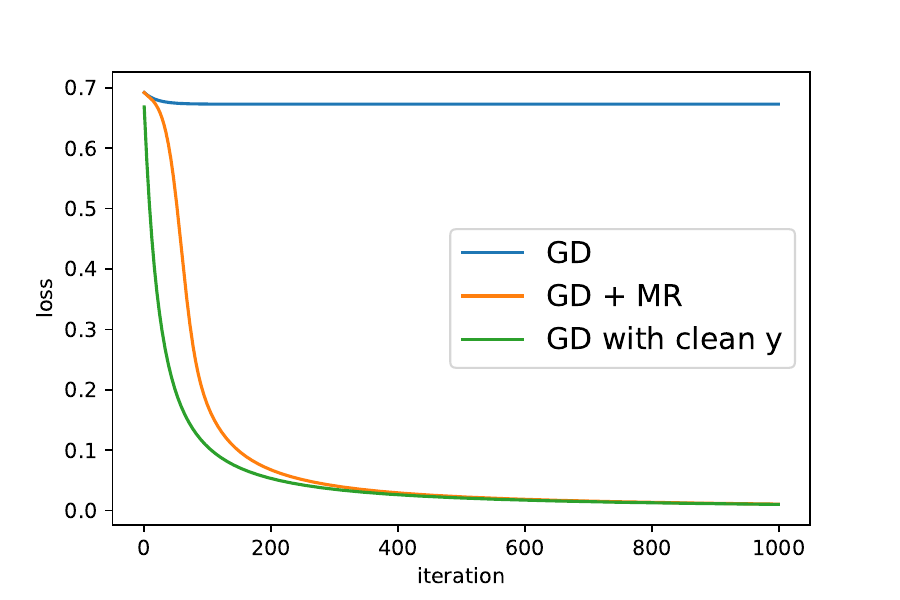}
    \caption{Logistic regression with GD.}
    \label{fig:short-a}
  \end{subfigure}
    \begin{subfigure}{0.32\linewidth}
    \includegraphics[width=1\linewidth]{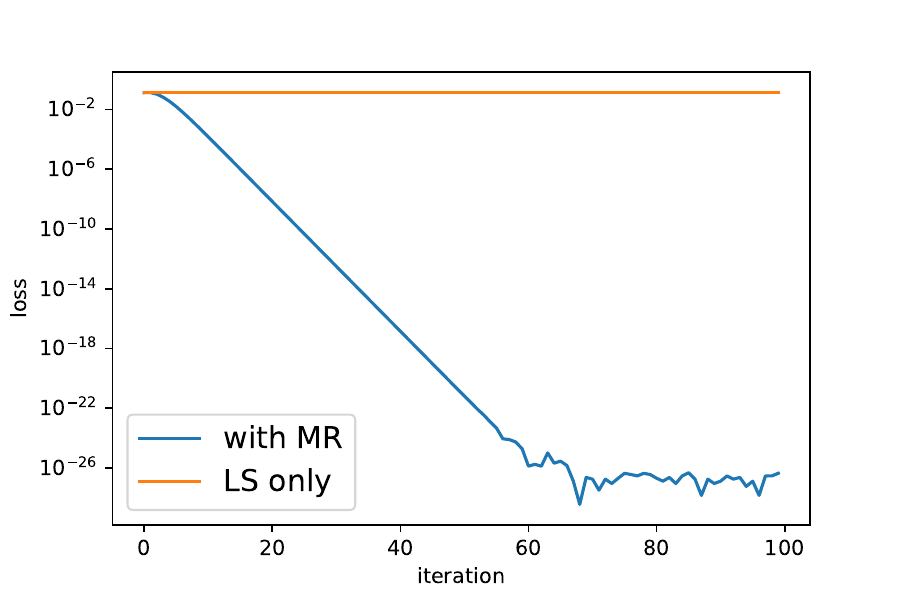}
    \caption{Linear regression with LS.}
    \end{subfigure}
    \begin{subfigure}{0.32\linewidth}
    \includegraphics[width=1\linewidth]{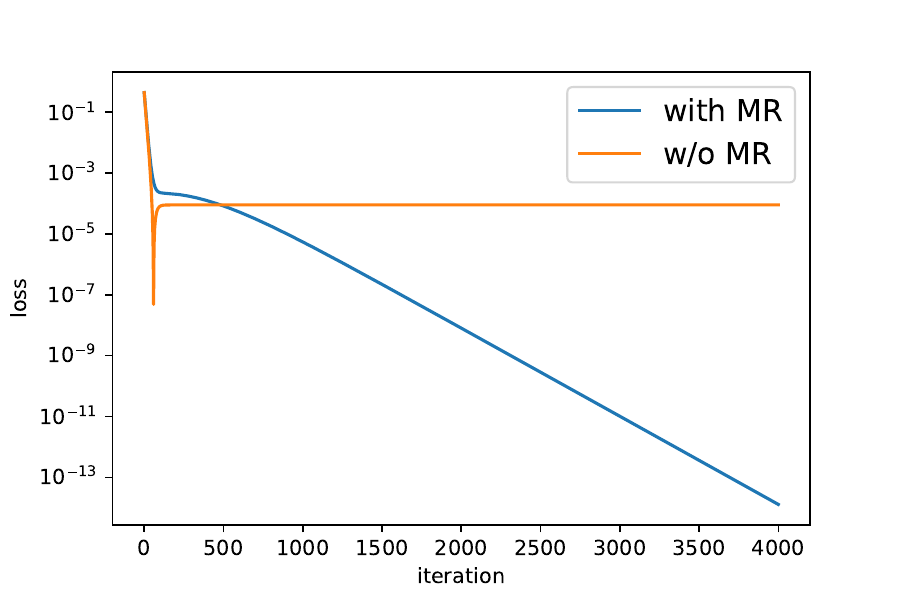}
    \caption{Linear regression with GD.}
    \end{subfigure}
  \caption{Loss evolution of the 1d illustrative examples with and without our MR technique.}
  \label{fig:illustrative}
\end{figure*} 

Convergence to a critical point does not necessarily implies that the point is good and robust solution.
To demonstrate the motivation of MR for noisy and corrupted labels, we present simple 1d binary examples of linear and logistic regression, where our method shows advantage theoretically and empirically.
Those examples provide a theoretical foundation for the benefit of using MR for robust training.
In this section, we demonstrate how, with MR, the majority of clean examples consistently receive higher weights and incur lower losses over time, contributing to improved results.
The weighting of training examples limits generalization ability. Generalization for unseen clean data is proportional to the number of clean training examples.

Consider the following setup. Let $x_i,y_i, \yT_i \in \mathbb{R}$, $1\leq i \leq N$, be the input samples and the clean and noisy labels. Denote by $\bs{x}$, $\bs{y}$, $\bs{\yT}$ the input data organized in vectors, $I_{cr} \subset [N]$ the set of indices where for $i\in I_{cr}$ it holds that $\yT_i \ne y_i$. 
In both examples $\abs{I_{cr}} = \sigma N$, where $\sigma = \frac{1}{2} - \Delta < \frac{1}{2}$ is the fraction of corrupted measurements and $\Delta$ is the distance of the corruption from $\frac{1}{2}$.
Let $L(\theta) = \frac{1}{N} \sum_{i=1}^N \ell(\theta;x_i, y_i)$ be the loss w.r.t clean $\bs{y}$ and $\theta_\infty$ the learned parameter when $t\rightarrow \infty$. We refer to $\bs{\yT}$ as the labels seen by the learner.

{\bf Logistic regression.} Our first examples is logistic regression for classification, where $\yT_i, y_i \in \{\pm1\}$, for $1 \leq i \leq N$, are the noisy and clean labels.
Let the inputs be $x_i \in \{\pm1\}$ such that the inputs are equal to the labels, $y_i = x_i$.
The noise we have in the labels is flip noise, i.e., $\forall k\in I_{cr}: \yT_k = -y_k$ and $\forall i \notin I_{cr}: \yT_i = y_i$.
We consider the non-stochastic case, both the input data, $\bs{x}$ and $\bs{y}$, are known and fixed and the full batched GD is deterministic.

When running GD with clean labels $\bs{y}$ we have $\theta_\infty\rightarrow\infty$, but running GD on the noisy version $\bs{\yT}$ yields convergence to finite $\theta_\infty$. Infinite $\theta$ is desirable when running logistic regression since the learner maximizes $\abs{x_i\theta}$ for having high confidence prediction.
Unlike GD, \cref{alg:gd_mw} does satisfy $\theta_\infty\rightarrow\infty$ as in the clean label case, which shows its advantage over using GD to overcome the noise. We now turn to show these results starting with clean GD.
 
\begin{lemma} \label{eq:logistic_clean_loss}
When running GD starting with $\theta_0 = 0$, using the clean labels $\bs{y}$  and step size $\alpha>0$, it holds that $\forall \epsilon > 0$ there exists $t > \frac{-\log(\exp(\epsilon) - 1)}{\alpha \exp(-\epsilon) (\exp(\epsilon) - 1)}$ such that $ L(\theta_t) = \log(1+\exp(-\theta_t)) < \epsilon$.
\end{lemma}
The above lemma implies that running GD with clean examples yields that the best solution is when $\theta_\infty \rightarrow \infty$ for minimizing the loss to 0.

While in clean logistic regression $\theta_t \rightarrow \infty$, the algorithm never converges to a finite parameter, it holds that in the noisy case, GD converges in finite time to a finite value.
\begin{lemma}\label{lemma:convergence_gd_logistic}
    When running GD with $\theta_0 = 0$, using the noisy labels $\bs{\yT}$ and step size $\alpha=1$, the algorithm converges to $\theta^\star = \log( \frac{1-\sigma}{\sigma})$. Formally,
    for any $\epsilon > 0$ and $T=O(\frac{1}{\epsilon})$ it holds that $\abs{\theta_t-\theta^\star} \leq \epsilon$.
\end{lemma}
The proof is simply done by finding a critical point of a $1$-smooth loss function (the loss is $1$-smooth since its second derivative is bounded by 1).
We use a step size of $\alpha=1=\frac{1}{\beta}$ since it guarantees an optimal convergence rate  (any smaller step size will result in lower convergence and larger may lead to an infinite loop around the critical point.).
Note that $\theta^\star = \theta_\infty$ is finite in contrast to the original clean problem. 
In this case, the learned parameter $\theta_\infty$ encounters $L(\theta_\infty) = c$ where $c$ is a strictly positive constant.

Now we will move to show the benefit of adding MW, which causes $\theta_\infty \rightarrow \infty$ as in the clean logistic problem.
To prove that, we first show that $\theta_t$ increases over time.
\begin{lemma}\label{lemma:increasing_w_logistic}
    When using \cref{alg:gd_mw} with step sizes $\eta=\alpha =1$, $\mu \geq 2$ and $\theta_0 = 0$, then $\theta_{t+1} > \theta_t$, $\forall t \geq 0$.
\end{lemma}
The proof follows from the rate of change in the weights assigned to clean and noisy examples, where clean examples receive higher weights.
This result allows us showing that \cref{alg:gd_mw} can reach any desirable small loss:
\begin{theorem} \label{theorem:gd_mw_logistic}
    When using \cref{alg:gd_mw} with step sizes $\eta=\alpha =1$, $\mu \geq 2$ and $\theta_0 = 0$, then for any
     $\epsilon>0$ exists $t \geq \max\Big\{ \frac{-\log(\frac{1}{2})}{\Delta} + 2, \frac{-\log(\exp(\epsilon)-1)}{\exp(-\epsilon)(\exp(\epsilon)-1)} \frac{2N}{\sigma} \Big\}$ such that $L(\theta_t) \leq \epsilon$.
\end{theorem}
The main claim in this theorem's proof is $\sum_{t=1}^\infty (\theta_{t+1} - \theta_{t}) \rightarrow \infty$, which yields that $\theta_\infty$ is unbounded. This results in an asymptotic decrease of the loss to 0.
The constrain $\mu > 2$ is due to $\sigma < \frac{1}{2}$ and when the noise ratio $\sigma$ is known the constrain can be relaxed to $\mu > \frac{1}{1-\sigma}$. 
Note that when $\Delta$  is very small (i.e the corruption ratio is approaching $\frac{1}{2}$) than it will take more optimization steps to reach a small loss, $\epsilon$.

This simple logistic regression example clearly shows the benefit of adding MW as the loss of GD alone converges to a solution with a non-zero loss while adding MW yields that the solution behaves like running GD with clean labels.

{\bf Linear regression.}
We turn to discuss a linear regression case, namely, minimizing the least squares (LS) loss of a linear model. We show the benefits of MR when combined not only with a GD learning procedure (as in \cref{alg:gd_mw}) but also with LS.
Recently, it was shown empirically that optimizing a classification DNN with LS is as good as learning it with the common log-loss \cite{muthukumar2020classification}. Furthermore, it is shown to provide an appealing implicit regularization \cite{poggio2020implicit}.

Here also $x_i = y_i \in \{\pm1\}$, but the noisy ``labels'' are $\bs{\yT} = \bs{y} + \bs{\epsilon}$, where $\epsilon_i = {\pm\epsilon}$ for $i\in I_{cr}$ and $\epsilon_i = 0$ for $i\notin I_{cr}$.
For example, if $x_1 = y_1 = -1$ and  $\epsilon_1 = 2$ then $\yT_1 = 1$, which corresponds to label flipping.
In this part, for a non-negative diagonal matrix $\bs{A}$, we use $\bs{\sqrt{A}}$ to denote applying an element-wise square-root on its entries.

As LS has a closed form solution for linear regression, we use it and compare to it instead of to GD.
For MR, our learning procedure alternates between updating a Weighted Least Square (WLS) solution and updating the weights as described in \cref{alg:ls_mw}. 
The loss this algorithm encounters at each step is weighted according to $p_t$ and thus, it uses LS solution of $\bs{x\sqrt{P_t}}$ and $\bs{\sqrt{P_t}\yT}$ to update $\theta_t$.
In this case, we do not limit the examples weights ($\mu=N$).

We show in the next theorem that in our setup, \cref{alg:ls_mw} manages to eliminate all noisy examples and converges to the optimum of the clean problem $\theta^\star = 1$.

\begin{theorem}\label{theorem:linear}
 For \cref{alg:ls_mw} we have that for $c >0 $ there exists $t > \frac{\ln(\frac{ \epsilon}{c+1+\Delta})}{\eta \epsilon^2 \Delta}$ such that $\abs{\theta_t - 1} \leq c$,
 where $\theta_t$ is the learned parameter by the algorithm.
\end{theorem}

First, to be able to discuss our results, we analyse $\theta_t$ and derive a simpler form of it than the one in \cref{alg:ls_mw}.
Let $\bs{\Sigma V_t}$ be the SVD decomposition of $\bs{x\sqrt{P_t}}$, where $\bs{\Sigma} \in \mathbb{R}^{1\cross N}$, $\bs{V_t}\in \mathbb{R}^{N\cross N}$ and $\bs{U}$ is omitted since $\bs{U} = 1$.
Note that the first vector in $\bs{V_t}$ is $\bs{x\sqrt{P_t}}$ (note that $ \norm{\bs{x\sqrt{P_t}}} = 1$) and the only non-zero singular value in $\bs{\Sigma}$ is $\sigma_1 = 1$. We also denote $\bs{\Sigma^2} = \bs{\Sigma^T\Sigma} \in\mathbb{R}^{N\cross N}$. With this notation, we have $\bs{x\sqrt{P_t}(\sqrt{P_t}x^T x\sqrt{P_t})^\dagger} = \bs{\Sigma V_t^T (V_t \Sigma^2 V_t^T)^\dagger}
     =  \bs{\Sigma V_t^T (V_t \Sigma^{\dagger} V_t^T)}
    = \bs{\Sigma V_t^T} = \bs{x \sqrt{P_t}},$
which leads to the solution $\theta_t = \bs{x \sqrt{P_t} \sqrt{P_t}} (\bs{y} + \bs{\epsilon}) = \bs{x P_t} (\bs{y} + \bs{\epsilon})$.

\cref{theorem:linear} proof relies on \cref{lemma:bounded_p} below, which lower bounds the probability of clean examples.
Ideally, in our 1d setting, a single clean label is enough for learning the optimal $\theta^\star$ with LS. Yet, the noise causes deviation from this solution. MR mitigates this problem, it assigns higher weights to clean data (as the lemma shows), which leads to a solution that is closer to the optimal one.
\begin{lemma}\label{lemma:bounded_p}
For any $t\in [T]$, $\forall i \notin I_{cr}$ it holds that $p_{t,i} \geq ((1-\sigma)N + \sigma N \exp (-\eta \epsilon^2 \Delta t))^{-1}$.
\end{lemma}
The lemma essentially states that the probability assigned to each clean example is increasing over time and goes to $((1-\sigma)N)^{-1}$ and the sum of all clean probabilities goes to $1$. 
{
\color{black} To demonstrate the effect of MR, we observe that after the first update of the learned parameter, the majority of clean examples drive the update toward reducing their losses. This effect intensifies over time, further reinforcing the focus on clean examples.
}
We now show how this lemma yields that the error in each iteration can be bounded with term decreasing over time.
Since $\theta_t=\bs{x P_t}(\bs{y} + \bs{\epsilon})$, as we have seen above, we have that the true loss at time $t$ for  $1 \leq i \leq N$ obeys
\begin{align*}
   \ell_{i}(\theta_t) &= \frac{1}{2} (x_i \theta_t - y_i)^2 = \frac{1}{2} (x_i\bs{x P_t}(\bs{y} + \bs{\epsilon}) - y_i)^2 \\ 
    & \hspace{-0.4in} = \frac{1}{2} (\bs{x P_t}\bs{y} - 1 + \bs{x P_t}\bs{\epsilon})^2 
    =\frac{1}{2} (\bs{x P_t \epsilon})^2
    \leq\frac{\epsilon^2}{2}\bigg( \sum_{j\in I_{cr}} p_{t,j}\bigg)^2
\end{align*}  
\cref{lemma:bounded_p} yields that the sum of corrupted probabilities $\sum_{j\in I_{cr}} p_{t,j}$ decreases over time since the probabilities for each clean example increases (the sum of probabilities equals 1).
Thus, the corrupted sum becomes closer to $0$, which yields that the loss with respect to the true labels obeys $L(\theta_t) \rightarrow 0$, and we converge to $0$ loss solution.

Note that the loss of the noisy LS solution with $\yT$ is $\frac{1}{2} \big(\sum_{j\in I_{cr}} x_j \epsilon_j\big)^2$. 
This implies that in many cases (e.g., label flipping), the LS solution yields a constant positive loss while our reweighting method results in a zero loss solution. When the corruption ratio approaches $\frac{1}{2}$ then more optimization steps are required to achieve low loss.

{\bf Theory demonstration.} We turn to demonstrate empirically the theoretical results above. 
We randomly generated $N = 1500$ examples, where the values of $x_i = y_i \in \{\pm1\}$ are chosen uniformly. We used $\epsilon = 1$ for the linear regression and $\sigma = 0.4$ for the corruption ratio in both cases.

The results in \cref{fig:illustrative} includes the logistic regression case and demonstrates clearly that the use of MR improves the learned parameter $\theta$ since the loss decays almost similarly to the loss when learning with clean $\bs{y}$.
The results illustrates \cref{lemma:convergence_gd_logistic}, \cref{lemma:increasing_w_logistic} and \cref{theorem:gd_mw_logistic}.
\cref{fig:illustrative} shows also an exponential decay of the loss when we learn a linear regression with MR, as stated in \cref{theorem:linear}.
We further simulate MR with GD (\cref{alg:gd_mw}) for linear regression, which we did not include in the theory above. We can observe the same convergence phenomena also there.

\begin{table*}[t]
\centering
\caption{Results on CIFAR10/100 for different label noise ratio. Note the advantage of using MR. } \label{tab:noisy_labels_cifar}
\begin{adjustbox}{width=0.85\linewidth}
\begin{tabular}{ cc| c c c c c c c c|c c c c c } 
 \toprule
Data                 & \shortstack{Noise \\ ratio}   & Base & \shortstack{Random\\ weighting} &\shortstack{Base+\\mixup}                       & \shortstack{Base+\\smoothing}                      &\shortstack{Base+\\mixup+\\smoothing}    & Arazo  & Co-teaching & \shortstack{Importance\\ reweighting}& MR         & \shortstack{\\ MR + \\mixup}& \shortstack{MR + \\ smoothing} & \shortstack{MR +\\mixup+\\ smoothing} \\
\midrule
\multirow{5}{*}{\small CIFAR10}& 0\%    & 95.15 & 93.07 &  95.97  & 95.15  & {\bf 96.26}   & 93.94 & 89.87 &  94.46     & 94.95       & 94.9            & 94.97 & 95.21\\
                        & 10\%   & 90.83 & 88.52 &  92.75  & 91.32  & 93.21         & 92.85      & 88.16 & 87.86 & 94.58       & 94.63           & 94.39 & {\bf 94.66}       \\
                        & 20\%   & 84.66 & 84.77 &  87.53  & 86.11  & 87.97         & 93.47       & 86.53 & 84.32 & 93.97       & 93.94           & 94.11 & {\bf 94.28}      \\
                        & 30\%   & 77.92 & 81,68 &  80.52  & 79.23  & 81.75         & {\bf 93.56} & 82.51 & 76.16 & 92.43       & 93.11           & 92.47 & 93.29  \\ 
                        & 40\%   & 69.22 & 78.77 &  72.29  & 71.08  & 83.87         & {\bf 92.7}  & 78.44 & 71.82 & 90.75       & 92.57           & 90.36 & 92.62      \\  
\midrule
\multirow{5}{*}{\small CIFAR100}& 0\%   & 74.77 & 72.74 & 78.00   & 76.09  & {\bf 79.36}   & 67.42 & 63.31 & 76.32 & 75.67 & 76.59           & 76.87 & 77.31           \\
                         & 10\%  & 63.59 & 66.97 & 72.35   & 69.95  & 74.42         & 70.77 & 62.23 & 70.72 & 67.04 & 73.92           & 74.42 & \textbf{76.13}     \\
                         & 20\%  & 57.08 & 58.58 & 66.38   & 63.72  & 70.5          & 70.43 & 61.47 & 63.41 & 62.62 & 72.41           & 70.5  & \textbf{73.9}      \\
                         & 30\%  & 52.12 & 50.95 & 62.23   & 56.16  & 64.43         & 69.71 & 57.83 & 55.28 & 56.37 & 70.4            & 67.11 & \textbf{71.4}\\
                         & 40\%  & 48.3  & 40.47 & 58.46   & 51.51  & 59.55         & 66.42 & 53.44 & 45.42 & 52.53 & 67.92           & 60.02 & \textbf{69.9}         \\
\bottomrule
\end{tabular}
\end{adjustbox}
\end{table*}

\begin{table}[t]
    \caption{Results on CIFAR10 (left) and CIFAR100 (right) with 40\% label noise, different optimizers and architecture. The best results within the same optimizer and architecture are in bold.}
\centering
    \begin{adjustbox}{width=.8\linewidth}
    \begin{tabular}{ccccc|cccc}
    \toprule
                                         & \multicolumn{2}{c}{Momentum} & \multicolumn{2}{c|}{Adam} & \multicolumn{2}{c}{Momentum} & \multicolumn{2}{c}{Adam}\\
                            Method      & WRN28-10        & ResNet18       & WRN28-10       & ResNet18 &  WRN28-10        & ResNet18       & WRN28-10       & ResNet18            \\ \midrule
     Base        & 75.58           & 69.22          & 63.84          & 72.4             & 55.3            & 48.3           &  45.63         & 37.34       \\
                             Base+MR     & {\bf 91.74}     & \textbf{90.75} & \textbf{79.04} & \textbf{88.32}   & \textbf{72.65}  & \textbf{52.53} & \textbf{58.42} & \textbf{61.68}   \\    
    \bottomrule  
    \end{tabular}    
    \end{adjustbox}
    \label{tab:arch_optim}
\end{table}

\begin{table}[t]
\centering
\caption{Comparison of MR to sparse regularization (SR) \cite{zhou2021learning}. CE stands for cross entropy loss.}
\label{tab:label_noise_sr}
\begin{tabular}{c|ccccc|ccccc}
\toprule
Method       & \multicolumn{5}{c|}{CIFAR10}                                              & \multicolumn{5}{c}{CIFAR100}                                                       \\
Noise Ratio  & 0\%            & 10\%            & 20\%           & 30\%           & 40\%          & 0\%            & 10\%           & 20\%           & 30\%           & 40\%           \\ \midrule
CE           & \textbf{90.27} & 82.48           & 74.46          & 68.14          & 58.8          & 70.44          & 61.42          & 54.49          & 40.8           & 39.61          \\
CE + SR      & 87.19          & 89.15           & 88.25          & 82.21          & 85.02         & \textbf{72.39} & 70.35          & 67.78          & 64.57          & 60.4           \\
CE + MR      & 90.14          & 88.95           & 84.88          & 81.4           & 76.38         & 69.34          & 64.65          & 59.36          & 46.46          & 44.41          \\
CE + SR + MR & 86.78          & \textbf{89.39}  & \textbf{88.44} & \textbf{82.65} & \textbf{85.1} & 71.8           & \textbf{71.05} & \textbf{69.75} & \textbf{67.11} & \textbf{64.11} \\ \bottomrule
\end{tabular}
\end{table}

\begin{table}[t]
\centering
\caption{Accuracy results on Clothing1M. $^*$for \cite{arazo2019unsupervised}, we take the number from their paper (same setup). SR is the method in \cite{zhou2021learning}}\label{tab:clothing1m}
    \begin{adjustbox}{width=1.\linewidth}
    \begin{tabular}{ccccccc|ccccc}
    \toprule
         Base & \shortstack{Random\\ Weighting} & \shortstack{Base +\\SR} & \shortstack{Base +\\mixup} & \shortstack{Base +\\ smoothing} & \shortstack{Base +\\mixup +\\  smoothing} &  Arazo & MR & \shortstack{MR +\\SR } & \shortstack{MR +\\ mixup} & \shortstack{MR +\\ smoothing} & \shortstack{MR +\\mixup +\\ smoothing} \\ \midrule
         68.94 & 69.96 & 70.17 & 70.22 & 69.04 & 70.65  & 71.0* & 70.69 & 70.35 & 69.98 & 71.12 & {\bf 71.18}  \\\bottomrule
    \end{tabular}
    \end{adjustbox} 
\end{table}

\begin{table}[t]
\centering
\caption{Accuracy results of MR with S2E \cite{yao2020searching}, UNICON \cite{karim2022unicon} and MW-net \cite{shu2019meta}.}
\label{tab:label_noise_s2e}
\begin{tabular}{c|ccc|ccc}
\toprule
              & \multicolumn{3}{c|}{CIFAR10} & \multicolumn{3}{c}{CIFAR100}                             \\
Noise Ratio  &  20\%          & 40\%   & 80\%       & 20\%            & 40\%     & 80\%      \\ \midrule
S2E \cite{yao2020searching}          &  60.13         & 55.51  &  \textbf{32.61}     & \bf{52.61}      & 46.29    & 16.41     \\
S2E \cite{yao2020searching} + MR     & \bf{60.6}      & \bf{56.59}  & 32.6 & 52.17           & \bf{48.22}  &  \textbf{17.79}   \\ \midrule
UNICON \cite{karim2022unicon}       & - & \bf{95.39}     & 93.25         & 78.05 & 77.13           & 63.97         \\
UNICON \cite{karim2022unicon} + MR    & - & 95.06          & \bf{93.32}    & \textbf{78.07} & \bf{77.56}      & \bf{64.54} \\ \midrule
MW-net \cite{shu2019meta} & 91.9 & 90.02 & 67.03 & 73.26 & 68.15 & \textbf{26.40}\\
MW-net \cite{shu2019meta} + MR & \textbf{92.42} & \textbf{90.1} & \textbf{69.28} & \textbf{73.44} & \textbf{69.07} & 22.26\\
\bottomrule
\end{tabular}
\end{table}

\begin{table}[t]
\caption{MR with Pair-flip and Instance dependent label noise.}\label{tab:diff_noise_types}
    \centering
    \begin{tabular}{cc|cc}
    \toprule
        Noise Type & Method & 20\% & 40\% \\ \midrule
Pair-flip & CE & 86.91 & 78.1 \\
        ~ & CE + MR & \textbf{88.85} & \textbf{78.51} \\ \midrule 
Instance dependent & CE & 68.27 & 50.45 \\ 
        ~ & CE + MR & \textbf{69.45} & \textbf{52.25} \\ 
\bottomrule
    \end{tabular}
\end{table}

\subsection{Lipschitzness of MR in the 1d case}
In this section we analyse the Lipschitzness of the MR solution in a 1d case.
Previously, a low Lipschitz constant w.r.t. the input data of a network was shown to improve adversarial robustness and hence it is used as a regularization for robust training \cite{jakubovitz2018improving, Ross2017InputGradients}.
In addition, low Lipschitz constant w.r.t. the parameters also yields robustness \cite{zhang2017mixup,Cisse17Parseval,Hein2017Formal}.
Specifically, a low Lipschitz constant yields high robustness against bounded $\ell_2$ norm attacks. 
This is due to the fact that the Lipschitz constant limits the change in the function (in $\ell_2$) when small perturbations are applied, and adversarial attacks are defined as bounded and unobservable perturbations to the input which change the output.

We focus on a 1d case with logistic loss, $\ell(x_i; \theta, y_i) = \log(1+\exp(-x_i\theta y_i))$, and $y_i\in \{\pm1\}$.
Since $G$-Lipschitzness is a general property of a function $\ell(x;y,\theta)$ that bounds the $l_2$ gradient norm w.r.t. the input $x$ over the whole domain, we focus on the effective weighted and average gradient $\sum_{i=1}^N p_i \abs{\frac{\partial \ell(x_i;y_i,\theta)}{\partial x_i}}$ for the MR case and $\frac{1}{N}\sum_{i=1}^N \abs{\frac{\partial \ell(x_i;y_i,\theta)}{\partial x_i}}$ for the GD case.
We present a lemma that analyzes \cref{alg:gd_mw} with a simpler MR update which does not encounter the history of the losses, i.e. $p_{i} = \frac{\exp(-\eta\ell(x_i; \theta, y_i))}{\sum_{j=1}^N \exp(-\eta\ell(x_j; \theta, y_j))}$.
\begin{lemma}\label{lemma:lipschitz}
    For an MW update of $p_{i} = \frac{\exp(-\eta\ell(x_i;y_i,\theta))}{\sum_{j=1}^N \exp(-\eta\ell(x_j;y_j,\theta))}$ (without history) and $y_i\in\{\pm1\}$ it holds that the uniform average of the absolute derivatives (GD) is larger than the MR weighting of absolute derivatives,
    \begin{align*}
        \frac{1}{N} \sum_{i=1}^N    \abs{\frac{\partial \ell(x_i;y_i,\theta)}{\partial x_i}} \geq \sum_{i=1}^N p_i \abs{\frac{\partial (x_i;y_i,\theta)}{\partial x_i}}.
    \end{align*}
\end{lemma}
The lemma states that using an update of the weights without the loss history yields a lower effective Lipschitzness in the MR case than in the uniform weighting case.

\begin{figure}[t] \label{fig:lipschitz}
    \centering
    \begin{subfigure}{0.49\linewidth}
    \includegraphics[width=1\linewidth]{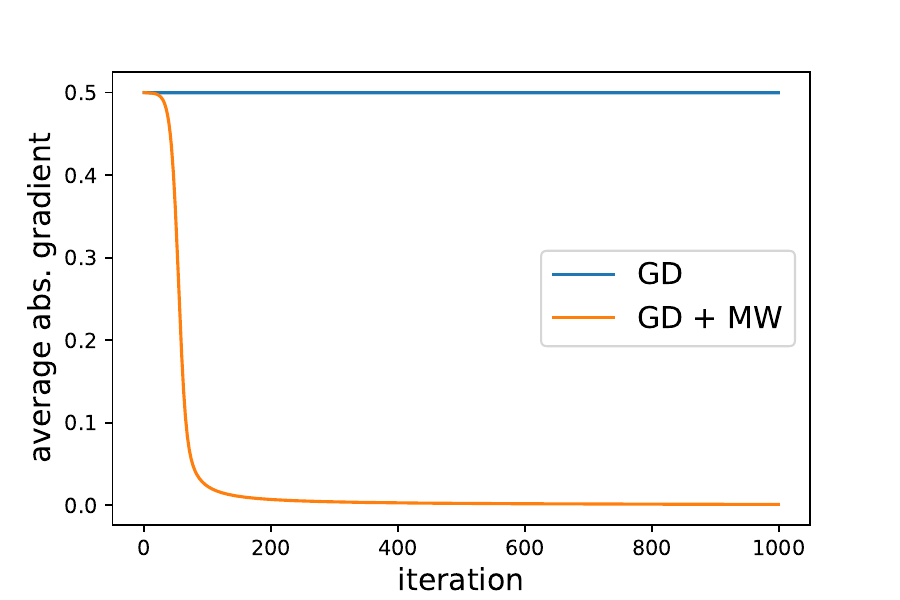}
    \caption{Effective absolute derivative.}
    \label{fig:lipschitz-a}
  \end{subfigure}
    \begin{subfigure}{0.49\linewidth}
    \includegraphics[width=1\linewidth]{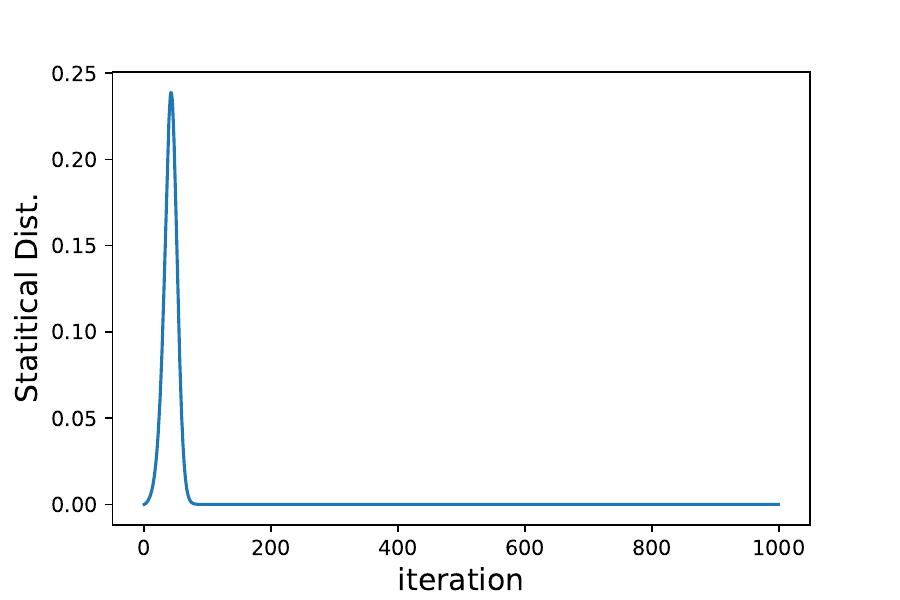}
    \caption{Statistical dist. $\nicefrac{1}{2}\norm{p_t-\Tilde{p}_t}_1$. \label{fig:lipschitz-b}}
    \end{subfigure}
  \caption{Simulations of Lipschitzness property of MR.}
\end{figure}

\textbf{Equivalence to full loss history.} We suggest an assumption that for large enough $t$ it holds that one can use an adjusted learning rate and a MW update without the history and get the same probabilities as MR with the losses history.
State the assumption formally, $p_{t,i} = \Tilde{p}_{t,i}$ where $\Tilde{p}_{t,i} = \frac{\exp(-\eta t \ell(x_i;y_i,\theta_t))}{\sum_{j=1}^N \exp(-\eta t\ell(x_j;y_j,\theta_t))}$ (note the step size, $\Tilde{\eta} = \eta t$).
According to the assumption, we could use \cref{lemma:lipschitz} and claim that MR solution has better Lipschitz property and is therefore more robust to adversarial attacks.
We demonstrate the validity of the assumption in \cref{fig:lipschitz}.

\textbf{Lipschitzness Simulations.} \label{sec:sim_lipschitz}
In order to demonstrate the results of the benefits of the Lipschitzness of MR we present a simulation.
We use $N=1500$ 1d input examples that are drawn randomly to be $x_i\in\{\pm1\}$ and $y_i\in\{\pm1\}$. 
We train GD and \cref{alg:gd_mw} with logistic loss.
In \cref{fig:lipschitz-a} we present the effective absolute gradient w.r.t. to the input and it is clearly shown that the MR has lower effective lipschitzness as \cref{lemma:lipschitz} suggests.
\cref{fig:lipschitz-b} plots the statistical distance between $p_{t,i} = \frac{\exp(-\eta\sum_{s=1}^t\ell(x_i;\theta_s, y_i))}{\sum_{j=1}^N \exp(-\eta \sum_{s=1}^t\ell(x_j;\theta_s, y_j))}$ and $\Tilde{p}_{t,i} = \frac{\exp(-\eta t \ell(x_i;\theta_t, y_i))}{\sum_{j=1}^N \exp(-\eta t \ell(x_j;\theta_t, y_j))}$, i.e. $\text{SD} = \frac{1}{2}\norm{p_t-\Tilde{p}_t}_1$.
In \cref{fig:lipschitz-b} it is clearly shown that our assumption is valid.

\section{Experiments}
\label{sec:exp}

\begin{figure}
    \centering 
    \includegraphics[width=0.48\linewidth]{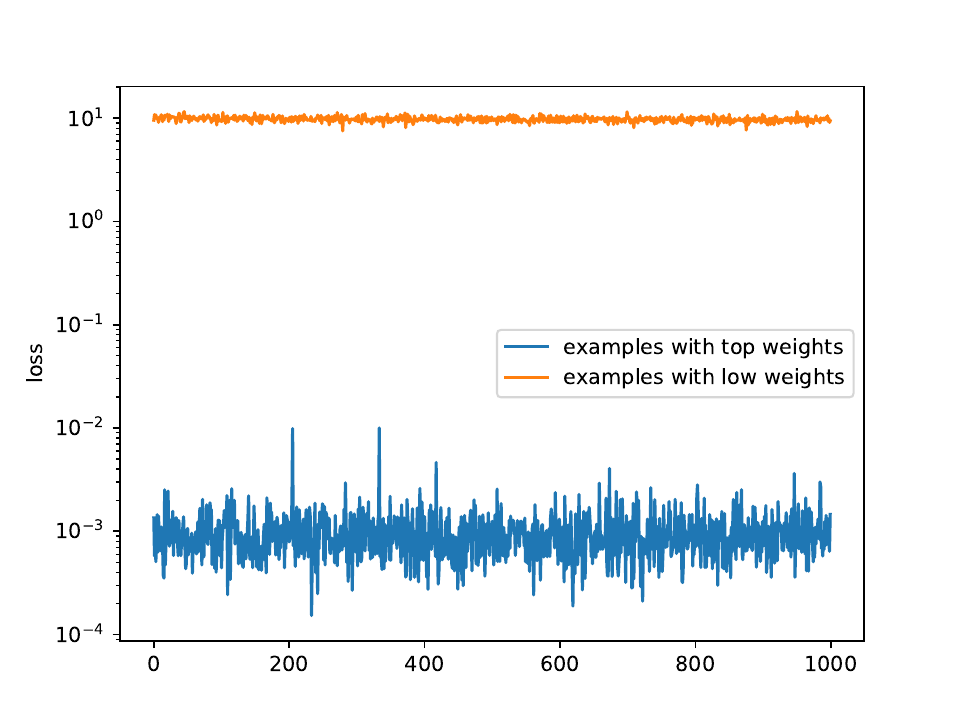}
    \includegraphics[width=0.48\linewidth]{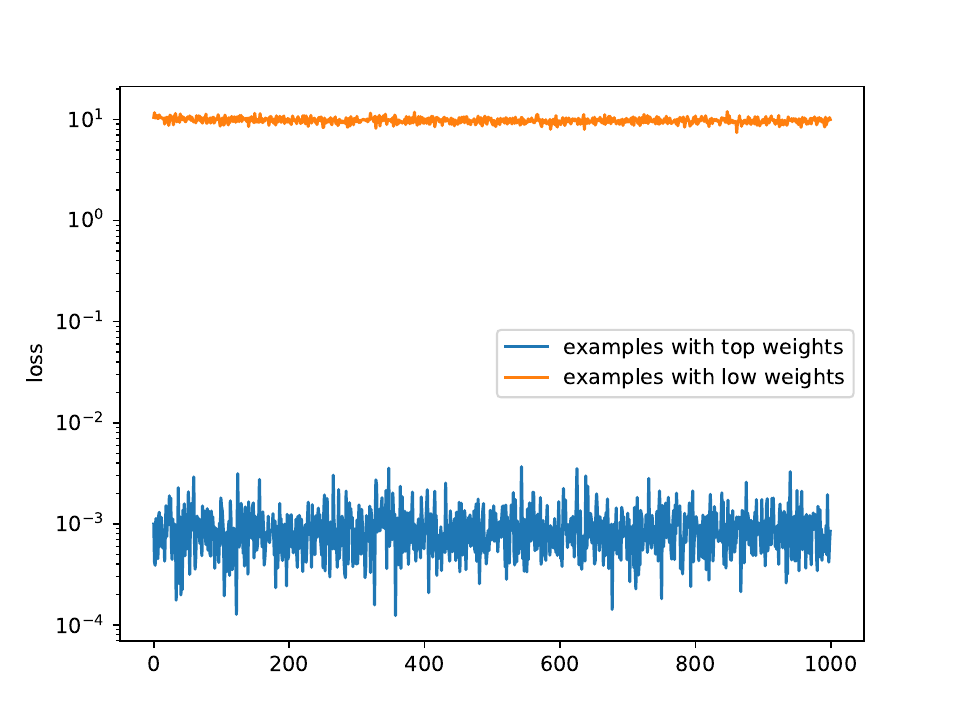}
  \caption{ Loss of examples with the $2\%$ highest and lowest weights. Left and right are training with 40\% and 20\% noise levels respectively.}
  \label{fig:top_low_losses}
\end{figure}

\subsection{Noisy Labels}
Motivated by the theoretical guarantees, we turn to show also empirically that MR improves robustness of DNNs.
We examine MR for training with noisy labels and in the presence of adversarial attacks at inference time.
We use artificial random label noise with CIFAR-10 and CIFAR-100 and natural label noise with the large-scale Clothing1M dataset \cite{xiao2015learning}.
For adversarial attacks, we add MR on top of well known robust training techniques and show improvement.

{\bf Artificial label noise.}
We start by evaluating MR on CIFAR-10 and CIFAR-100. 
To examine our method in the presence of label noise, we changed each label independently with a constant uniform probability to a random one.
Clearly, the change is done only in the training set while the test set is noise free.
We use ResNet-18 \cite{he2016deep} as the classification network and a constant MR step size $\eta = 0.01$, which is found by a grid search; the full search results appear in the \cref{fig:grid_search}. 
We train the network for 200 epochs. An initial learning rate of $\alpha = 0.1$ is used and then multiplied by $0.1$ at epochs 80 and 120.
We use momentum with parameter $0.9$, $l_2$ regularization with factor of $5\times 10^{-3}$ and $10^{-3}$ for CIFAR-10 and CIFAR-100 respectively, random crop and random horizontal flip.

The $\mu/N$ upper bound of the weights in MR is necessary to theoretically ensure that MR does not converge to degenerated $p$ vectors.
We examine the effect of $\mu$ on the accuracy and the distribution $p$. In the \cref{fig:upper_bound} we experiment with multiple values and observe that high $\mu$ values that do not limit the weighting do not affect the performance.
Hence, in order to avoid the worst-case complexity of $O(N^2)$, we use unlimited probabilities (i.e. $\mu=N$), where we simply use $l_1$ normalization ($O(N)$ complexity).
Thus, we can infer that practically MR implicitly exploits the large training set and the non-degenerate loss distribution which the DNN produces, and without a need for additional constrains it does not convergence to undesirable $p$.

We compare and combine our method with label smoothing \cite{szegedy2016rethinking} and mixup \cite{zhang2017mixup}. 
The label smoothing parameter is set to $0.1$ and in mixup to $\alpha=1$.
These baselines are common tools for improving robustness of networks and specifically we use them since they do not need an additional separate set of clean training examples as is required in many techniques for training with noisy labels.
We also conduct experiments with WRN28-10 \cite{zagoruyko2016wide} and Adam optimizer \cite{kingma2015adam} with learning rate $0.001$.
In addition we compare our method to random weighting (also appear in \cite{ren2018learning}), where the weights are randomly distributed according to rectified gaussian distribution, i.e,  $w_{t,i}^{rnd} = \frac{\max(0, z_{t,i})}{\sum_j z_{t,j}}, \quad z_i\sim \mathcal{N}(0,1)$.

We use as baseline, the unsupervised weighting scheme in \cite{arazo2019unsupervised}, which also does not require an additional small clean train set.
We employ the ``M-DYR-H'' variant with the hyper-parameters as in their paper. This variant includes mixup, dynamic and hard bootstrapping.
We also use Importance reweighting \cite{Liu16Classification} and Co-teaching \cite{han2018co} as baselines.

\Cref{tab:noisy_labels_cifar} summarizes the results and shows the superiority of using MR with respect to the SGD (Base) in the presence of label noise and that MR only slightly decreases accuracy for clean labels.
Note that MR improves the accuracy when used with mixup and the training contains noisy labels.
This phenomena is observed also with label smoothing and when using both label smoothing and mixup.
Adding MR improves the results more significantly as the noise is more severe and the classification is harder, which is the case in CIFAR-100.
In the majority of the scenarios, MR also outperforms \cite{arazo2019unsupervised}, especially on CIFAR-100.
In the supplementary material attached to this paper, we provide an analysis of the statistical significance of the improvement achieved by incorporating MR.

The results with WRN28-10 and Adam optimizer for 40\% label noise are found in \cref{tab:arch_optim}.
As can been seen, the MR advantage remains when changing the architecture or the optimizer.
Additional results with Adam, MR, mixup, label smoothing and methods which requires clean data \cite{shu2019meta,hendrycks2018using} appear in the \cref{tab:add_baselines}. 
Overall, we can say that MR is oblivious to changes in the optimizer and architecture, and it enhances the performance independently of those variations when used for label noise.

We compare our strategy also to the recent state-of-the-art (SOTA) sparse regularization (SR) method \cite{zhou2021learning}.
For this comparison we use the settings detailed in their paper, i.e., 8-layers CNN for CIFAR-10 and ResNet34 for CIFAR-100. Additional details appear in supp. mat.
\cref{tab:label_noise_sr} exhibits the performance of MR with SR. It demonstrates again the improvement provided by MR in the presence of label noise when added on top of another algorithm.

Since MR can also be viewed as a curriculum learning method we compare it with S2E \cite{yao2020searching} which changes the sampling probabilities from which the mini-batches are sampled. Additionally, we compare to a method named UNICON \cite{karim2022unicon} and MW-net \cite{shu2019meta}. For all of the methods the results are reported according to the settings described in their original papers.
As for \cite{karim2022unicon}, when combined with MR, we use MR only for the warmup stage, $10$ and $30$ epochs for CIFAR10 and CIFAR100 respectively.
No results are reported for UNICON with 20\% noise level with CIFAR-10 since the results are not reproducible with their official code. 
Note that MW-net requires additional clean set of examples. We employ MR only on the part of MW-net that uses the noisy train set.
Additional details appear in the supp. mat. 
In \cref{tab:label_noise_s2e} it is shown that in most cases the addition of MR on top of the other methods improves the accuracy.

{\color{black}
Finally, we compare our method to the semi-supervised SOTA method DivideMix, which employs additional losses and trains two networks. In \cref{tab:time}, we report accuracy results on CIFAR-10 and observe that although DivideMix achieves higher accuracy than our MR, it does so at the cost of significantly longer training time.
}

\begin{table*}[t]
\centering
\caption{Robustness against adversarial attacks. Note the improvement of MR when added to adv. training, and combined with mixup.}
\label{tab:adv_rob_results}
\begin{adjustbox}{width=1.\textwidth}
\begin{tabular}{ lc c c c cc c|c c c c c c c} 
\toprule
                    & \multicolumn{7}{c|}{CIFAR10}                                                   & \multicolumn{7}{c}{CIFAR100} \\\midrule
Method              & Natural Images& PGD           & PGD          & PGD         & FGSM      & FGSM       & FGSM        & Natural Images& PGD           & PGD          & PGD        & FGSM          & FGSM      & FGSM       \\
$\epsilon$          & $0$           & $0.01$        & $0.02$       & $8/255$     & $0.01$    & $0.02$     & $8/255$     & $0$           & $0.01$        & $0.02$       & $8/255$    & $0.01$        & $0.02$    & $8/255$        \\\midrule
Base                & 94.02         & 25.62         & 5.1          & 3.6         & 72.48     & 51.88      & 41.83       & 71.31         & 21.15         & 10.4         & 9.39       & 34.54         & 21.77     & 15.79      \\
Random Weight.      & 92.57         & 24.02         & 5.76         & 5.2         & 47.18     & 27.04      & 20.41       & 71.15         & 20.66         & 11.12        & 10.07      & 33.16         & 21.06     & 16.4           \\
Base+MR             & 94.71         & 29.27         & 5.47         & 3.76        & 74.81     & 51,4       & 44.72       & 71.95         & 23.02         & 11.00        & 9.44       & 37.58         & 24.51     & 18.85      \\
Base+mixup          & {\bf 95.95}   & 24.89         & 3.23         & 2.46        & 77.38     & 71.96      & 68.18       & {\bf 77.91}   & 14.59         & 8.4          & 7.95       & 39.5          & 32.36     & 29.26      \\ 
Base+mixup + MR     & 95.5          & 8.33          & 3.04         & 3.12        & 58.42     & 44.98      & 39.88       & 75.86         & 12.60         & 8.67         & 8.37       & 39.09         & 32.20     & 29.24      \\\midrule
         
Free \cite{shafahi2019adversarial}                & 90.32         & 81.11         & 64.28        & 39.09       & 83.22     & 73.8       & 57.94       & 61.46         & 52.43         & 38.81        & 26.55      & 53.74         & 43.93      & 35.15     \\
Free \cite{shafahi2019adversarial}+Random weight. & 89.18         & 79.85         & 61.68        & 39.53       & 81.24     & 69.13      & 56.28       & 60.26         & 50.76         & 36.86        & 24.92      & 52.03         & 41.44      & 32.12    \\
Free \cite{shafahi2019adversarial}+MR             & 90.62         & 82.6          & 65.37        & 40.24       & 84.25     & 73.45      & 60.02       & 62.75         & 54.19         & 40.84        & 29.0       & {\bf 55.64}   & 46.08      & 37.7     \\
Free \cite{shafahi2019adversarial}+mixup          & 89.54         & 81.77         & 71.19  & 44.54       & 83.29     & {\bf 77.92}& 63.71       & 60.6          & 52.30         & 39.27        & 27.76      & 53.69         & 44.65      & 37.44     \\
Free \cite{shafahi2019adversarial}+MR+mixup       & 89.4          & {\bf 83.49}   & 69.72        & 46.66 & {\bf 84.47}& 76.63     & 67.56 & 59.74         & {\bf 54.59}   & 44.45  & 33.41 & 55.19         & {\bf 47.89}& 40.2   \\ \midrule
TRADES \cite{zhang2019theoretically}     & 84.88   & 81.4    & 72.66       & 59.72      &81.69 & 75.31      & 67.83      & 55.69         & 51.98         & 45.14    & 36.59        & 52.27         & 46.46       & 40.04 \\
TRADES \cite{zhang2019theoretically} + MR& 83.4          & 81.28         & {\bf 73.47} & {\bf 60.76}& 81.53     &  75.72 & {\bf 68.74}& 56.32   & 52.47    & {\bf 47.12}& {\bf 37.45}& 52.76   & 47.12 & {\bf 40.86} \\\bottomrule
\end{tabular}
\end{adjustbox}
\end{table*}

\begin{table}[t]
\centering

\caption{Robustness against adversarial attacks with WideResNet 34-10. All attacks are limited to $\epsilon=8/255$.}
\label{tab:adv_rob_results_new}
\begin{tabular}{ lc c c c cc} 
\toprule
Method              & Natural Images& PGD           & PGD          & PGD         & CW      & AA              \\
Iterations          & -           & $10$        & $20$       & $50$     & $20$    & -         \\\midrule
LAS-AT \cite{jia2022adversarial}& \textbf{86.95} & 56.27 & 55.02      & 54.81    & 54.98   & 52.7                     \\
LAS-AT \cite{jia2022adversarial} + MR& 83.97 & \textbf{57.2} & \textbf{56.31}       & \textbf{56.06}    & \textbf{56.0}  & \textbf{54.39}                       \\ \midrule
MAIL-TRADES \cite{liu2021probabilistic}& \textbf{79.35} & 43.81 & 42.24 &  41.65   & 40.81    & 38.51                       \\
MAIL-TRADES \cite{liu2021probabilistic} + MR& 75.26 & \textbf{49.25} & \textbf{48.44} & \textbf{48.28}       & \textbf{46.01}   & \textbf{45.11}                         \\
\bottomrule
\end{tabular}
\end{table}

{\bf Pair-flip and Instance-dependent label noise. }
We applied MR in the presence of artificial yet more realistic label noise. Specifically, we employed pair-flip label noise \cite{han2018co}, in which related classes are flipped, and instance-dependent label noise \cite{xia2020part}, where the labels are altered based on the image's parts.
The accuracy results for these two types and levels of noise are presented in \cref{tab:diff_noise_types}. In all the tested cases, the utilization of MR consistently improves performance.

{\bf Realistic noisy labels.}
We examine MR on the Clothing1M \cite{xiao2015learning} dataset which includes 1M images of clothing products labeled by their keywords.
The dataset includes $14$ different classes and is known to be noisy.
We used the same hyper-parameters as in Meta-Weight \cite{shu2019meta}, with ResNet-50 \cite{he2016deep} pre-trained on ImageNet \cite{deng2009imagenet} and we train it for additional 10 epochs.
We used $\eta=0.01$ for MR step size and multiply it by $1.5$ after $2$ and $6$ epochs. 
Instead of using MR after each epoch we employ it after half an epoch since  we found that $10$ weighting updates are not significant enough.

\cref{tab:clothing1m} details the results and shows clearly the improvement of MR over SGD with $\sim 1.5\%$ enhancement.
Note that adding MR improves the results when used with the base optimizer, label smoothing and with the combination of mixup and label smoothing.
Using SR \cite{zhou2021learning} with MR is better than employing SR alone.
Employing MR, mixup and label smoothing leads to a significant improvement of $\sim$$2$\% over SGD and even outperforms \cite{arazo2019unsupervised}.
Thus, we find that MR is useful and favourable also with real label noise.

\subsection{Adversarial Robustness}
Another type of harmful noise to DNN is the one used in adversarial attacks. 
 It had been shown that small perturbations to the input data (during inference) that are unrecognized visually can lead to an entirely different outcome in the network's output~\cite{szegedy2014intriguing}.
 A notable leading strategy among the defense methods is adversarial training \cite{Goodfellow15Explaining,madry2018towards} that adds adversarial examples in the training of the DNN to improve its robustness.

We have established a connection between MR and robustness to adversarial attacks using the relationship we show to low Lipschitz constant in \cref{sec:sim_lipschitz}. In addition, an intuitive explanation about the connection between the two can be that relying on examples with high confidence (low loss) during optimization can lead to improved adversarial robustness \cite{liu2021probabilistic,zhang2020geometry}.

 To empirically assess the benefits of MR in this setting, we employ it with Free Adversarial Training \cite{shafahi2019adversarial} and TRADES~\cite{zhang2019theoretically}, where both are well established and efficient methods for robust training.
We focus on them due to their efficiency in resources.
We evaluate MR with non-corrupted images (e.g. ``Natural Images'') and against adversarial examples produced by the PGD \cite{madry2018towards} and FGSM \cite{Goodfellow15Explaining} attacks with a bounded $\ell_\infty$ norm of $\epsilon \in \{0.01 , 0.02, 8/255\}$.
All experiments setting as well as additional tests (e.g., MR with Adam) appear in \cref{tab:adv_adam}.

\Cref{tab:adv_rob_results} includes the results for base, mixup and robust training baselines (i.e. Free \cite{shafahi2019adversarial} and TRADES \cite{zhang2019theoretically}) with and without MR.
As can been seen from the results, adding MR weighting on top of robust training yields improvement most scenarios tested and is more significant with CIFAR-100.
Adding MR to free training is more beneficial than adding mixup for CIFAR-100 while in CIFAR-10 the results are indefinite.
As for base training, MR slightly improve robustness, yet the accuracy against strong attacks is poor.
In addition, the accuracy of base with both mixup and MR in all scenarios is worse than at least one of the methods alone.
When our method is added on top of the combination of robust training and mixup we can see improvement against all attacks except the ones with CIFAR-10 and $\epsilon=0.02$, but we observe a slight degradation in accuracy with natural images.
In most attacks the combinations with MR result in superior accuracy against adversarial examples. 

Note that free training and its combination with MR and mixup do not perform as well as the regular networks on the natural data (without attacks). This problems also occurs  with other robust training methods, but can be mitigated by using additional unlabeled data in the training \cite{carmon2019unlabeled,zhai2019adversarially}.

We also apply MR with the recent adversarial training method LAS-AT \cite{jia2022adversarial} and MAIL \cite{liu2021probabilistic}.
We use WRN 34-10 models and attacks that are bounded according to $l_\infty$ with $\epsilon=8/255$ but differ in the number of iterations.
It is shown in \cref{tab:adv_rob_results_new} that using MR on top of this adversarial trainings improves the robustness of the networks. As seen also in some of the other baselines, accuracy with natural images decreases when MR is employed.

{\color{black}
\subsection{\textbf{Training Time}} Our method incurs additional training time due to the computation of the weights. In \cref{tab:time}, we report the running times of Arazo \cite{arazo2019unsupervised} and DivideMix \cite{li2020dividemix}. DevideMix is a semi-supervised baseline that requires extra time to compute augmentations, losses, and to perform additional forward and backward passes. While our accuracy is lower than that of a SOTA semi-supervised method, this method requires more than seven times the training time compared to MR. This demonstrates that our approach offers a more efficient alternative when training time is a key consideration.

\begin{table}[h]
\centering
\begin{tabular}{l|ccc|cc}
\toprule
 & \multicolumn{3}{c|}{Time} & \multicolumn{2}{c}{Accuracy} \\
 & Epoch Time\tiny{[sec]} & \# Epochs & Total\tiny{[min]} & 20\% & 40\% \\
\midrule
DivideMix & 88.38 & 300 & 441.9 & \textbf{96.15} & \textbf{95.41} \\
Arazo & \textbf{18.75} & 300 & 93.75 & 93.47 & 92.7 \\
MR & 30.31 & \textbf{120} & \textbf{60.62} & 94.28 & 92.62 \\
\bottomrule
\end{tabular}
\caption{{\color{black} Training time and accuracy on CIFAR-10. All training times were measured on the same machine and under identical conditions, using the same batch sizes. The number of training epochs follows that reported in the original papers. Note that MR achieves significantly shorter training time compared to DivideMix, while DivideMix reaches higher accuracy. }}\label{tab:time}
\end{table}

}

\section{Conclusion}
This work proposed the MR method, which is  motivated by learning with expert advice and
use the MW algorithm for reweighting examples along with the common SGD DNN optimization. 
We proved theoretically the benefits of using this reweighting strategy for DNN training and showed through various experiments that MR improves DNN robustness for learning with label noise.
We provided also evidence that MR enhances robustness also to adversarial examples showing improvement of leading adversarial approaches.
One concern of the use of MR is training with class imbalance which may lead to assigning low weights to the examples with minority class since they suffer high loss.
Future work will involve finding additional tools inspired by learning with expert advice that are appropriate for the imbalanced scenario.
We further suggest investigating the effect of using variants of MW such as Follow the Regularized Leader and Online Mirror Descent \cite{shalev2011online} which imply different senses of decreasing the MW step size and may exhibit other robustness properties.
Another direction is theoretically investigating the generalization abilities of our method.

To summarize, we believe that the results shown in this work position MR as a useful techniques in the DNN toolbox that can elevate its robustness along with other tools like mixup and label smoothing.

\section*{Acknowledgments}
We thank Gilad Cohen for the adversarial training code. This research has received funding from the European Research Council (ERC) under the European Union’s Horizon 2020 research and innovation programme (grant agreement No. 101078075), and is also supported by ERC-StG SPADE (PI Giryes), the Israel Science Foundation (ISF, grant numbers 2549/19 and 3174/23), the Lenç and the Blavatnik Family Foundation, the Yandex Initiative in Machine Learning, and the Tel Aviv University Center for AI and Data Science (TAD).

Views and opinions expressed are however those of the author(s) only and do not necessarily reflect those of the European Union or the European Research Council. Neither the European Union nor the granting authority can be held responsible for them.

\appendix

\section{Projection Algorithm} \label{sup:projection}
In \cref{alg:projection} we detail the algorithm of efficient projection to a constrained domain, where each probability has limited value of $\mu/N$. This algorithm was introduced in prior works in \cite{warmuth2008randomized,koolen2010hedging}. 
The algorithm normalizes the weights into a probability and then iteratively distributes all the mass that exceeds the upper bound.
The computation complexity is $O(N^2)$, where $N$ is the number of examples.
\begin{algorithm}[H]
   \caption{KL-projection to constrained domain \cite{warmuth2008randomized,koolen2010hedging}}\label{alg:projection}
\begin{algorithmic}
   \STATE {\bfseries Input:} unconstrained weighting $w_{i}$, maximal weighting ratio - $\mu$.
   \STATE {\bfseries Output:} distribution $p\in\Delta_N$ s.t.~$p_i < \frac{\mu}{N},~ \forall i = 1,...,N$.
   \STATE \textbf{Normalize:} $p_{i} = \frac{w_{i}}{\sum_{j=1}^N w_{j}}$
    \WHILE{$\max_{i=1,...,N} p_i > \frac{\mu}{N}$}
    \STATE \textbf{Excess mass:} $r = \sum_{i=1}^N \max\{0, p_i - \frac{\mu}{N}\}$
    \STATE \textbf{Redistributing:} update $p_i$ s.t.~${p_i < \frac{\mu}{N}}$: \\
    ${p_i \gets p_i + r\cdot\frac{p_i}{\sum_{j, p_j < \frac{\mu}{N}} p_j}}$
    \ENDWHILE
\end{algorithmic}
\end{algorithm}

\section{Proofs} \label{sec:appendix}

Here we present the proofs of the lemmas and theorems presented in the paper. For the convenience of the reader, we repeat the definitions of the lemmas and theorems that we prove.

\subsection{Relevant Definitions} \label{sup:definition}
Prior to our proof we detail the definitions of the terms we use.
\begin{definition}
A function $f: W \rightarrow \mathbb{R}$ is $B$ bounded if:
\begin{align*}
    \forall u \in W: f(u) \leq B.
\end{align*}
\end{definition}

\begin{definition}
A function $f: W \rightarrow \mathbb{R}$ is $G$-Lipschitz ($G>0$) if:
\begin{align*}
    \forall u,v \in W: \norm{f(u) - f(v)} \leq G \norm{u-v}.
\end{align*}
\end{definition}

\begin{definition}
A differentiable function $f: W \rightarrow \mathbb{R}$ is $\beta$-smooth if:
\begin{align*}
    \forall u,v \in W: \norm{\nabla f(u) - \nabla f(v)} \leq \beta \norm{u-v}.
\end{align*}
\end{definition}

\subsection{Proof of \cref{lemma:descent_lemma}}
% \begin{replemma}{lemma:descent_lemma}
\begin{lemma}
For a $\beta$-smooth loss $\ell(\cdot)$, and $\theta_{t+1}$, $p_{t+1}$ updated as in \cref{alg:gd_mw} with GD step size of $\alpha=\frac{1}{\beta}$ and MW step size $\eta > 0$, we have:
\begin{align*}
    \sum_{i=1}^N \left( p_{t+1, i} \ell_i(\theta_{t+1}) -  p_{t, i}  \ell_i(\theta_{t}) \right)
    \leq -\frac{1}{2\beta} \norm{\sum_{i=1}^N p_{i,t} \nabla \ell_i(\theta_t)}^2&.
\end{align*}
% \end{replemma}
\end{lemma}
\begin{proof}
Note that $\sum_{i=1}^N p_{t, i} \ell_i(\theta)$ is $\beta$-smooth in $\theta$. Using $\beta$-smoothness and the GD step definition in \cref{alg:gd_mw},  we get:
\begin{align*}
    \sum_{i=1}^N p_{t, i} \ell_i(\theta_{t+1}) 
    &\leq \sum_{i=1}^N p_{t, i} \ell_i(\theta_{t}) 
     - \alpha \norm{\sum_{i=1}^N p_{t, i} \nabla \ell_i(\theta_t)}^2 
    \\&+ \frac{\alpha^2\beta}{2}  \norm{\sum_{i=1}^N p_{t, i} \nabla \ell_i(\theta_t)}^2 .
\end{align*}

Since $\alpha = \frac{1}{\beta}$, we get:
\begin{align*}
    \sum_{i=1}^N p_{t, i} \ell_i(\theta_{t+1}) \leq \sum_{i=1}^N p_{t, i} \ell_i(\theta_{t}) 
    - \frac{1}{2\beta} \norm{\sum_{i=1}^N p_{t, i} \nabla \ell_i(\theta_t)}^2 .
\end{align*}

All that is left to be shown is that:
\begin{align} \label{optimality_p}
    \sum_{i=1}^N p_{t+1, i} \ell_i(\theta_{t+1}) \leq \sum_{i=1}^N p_{t, i} \ell_i(\theta_{t+1}) .
\end{align}

All examples are updated using an MW step followed by a projection. 
Thus, $p$ updates can be interpreted as FTRL (Follow The Regularized Leader) algorithm with a constant step size $\eta$ \cite{amir2020prediction} and the probability update can be written formally as
\begin{align*}
    p_{t+1} = \argmin_{\genfrac{}{}{0pt}{}{\theta, w\in \Delta_N}{w_i < \mu/N},} \Big\{ w \cdot \sum_{s=1}^{t+1}  \ell(\theta_s) + \frac{1}{\eta} \sum_{i=1}^N w_i \ln(w_i) - w_i\Big\},
\end{align*}
where $\Delta_{N}$ is $N$-simplex, $p_{t+1}\in\Delta_N$ is a vector and $\ell(\theta_s)$ are treated as known fixed vectors.

Define a function $R$ to be $R(w) = \sum_{i=1}^N w_i \ln(w_i) - w_i$.
Let $\Phi(u) = \sum_{i=1}^N u_i \sum_{s=1}^t l_i(\theta_s) + \frac{1}{\eta}R(u)$ and $\phi(u) = \sum_{i=1}^N u_i l_i(\theta_{t+1})$.
$\Phi$ is convex since it is a sum of linear functions and a negative Shannon entropy which are both convex. Note that $\phi$ is also convex since it is linear. Using the convexity of $\Phi$ and the fact that $p_t$ is its minimum, we have that
\begin{align} \label{eq_phi}
\Phi(p_{t+1}) - \Phi(p_t) \geq 0 .
\end{align}
From the minimality of $p_{t+1}$ w.r.t $\Phi + \phi$, we have
\begin{align} \label{eq:Phi_phi}
(\Phi + \phi)(p_{t}) - (\Phi + \phi)(p_{t+1}) \geq 0 .
\end{align}
Following \cref{eq_phi} and \cref{eq:Phi_phi}, we get
\begin{align*}
    \phi(p_{t}) - \phi(p_{t+1}) \geq \Phi(p_{t+1}) - \Phi(p_t) \geq 0 .
\end{align*}
Plugging the definitions of $\phi$ and $\Phi$ leads to \cref{optimality_p}, which completes the proof.
\end{proof}

\subsection{Proof of \cref{thm:gd_convergence}}
\begin{theorem}
For a $\beta$-smooth loss $\ell(\cdot)$, and $\theta_{t+1}$, $p_{t+1}$ being updated as in \cref{alg:gd_mw} with GD step size of $\alpha=\frac{1}{\beta}$ and MR step size $\eta > 0$, we have that
\begin{align*}
    \frac{1}{T} \sum_{t=1}^T &\norm{\sum_{i=1}^N p_{t,i} \nabla \ell_i(\theta_t)}^2 
    \leq \\
    & \quad \frac{2\beta}{T} \bigg(\frac{1}{N}\sum_{i=1}^N \ell_i(\theta_0) - \sum_{i=1}^N p_i^\star \ell_i(\theta^\star)\bigg),
\end{align*}
where $p^\star, \theta^\star = \argmin_{\genfrac{}{}{0pt}{}{\theta, p\in \Delta_N}{p_i < \mu/N},}  \sum_{i=1}^N p_i \ell_i(\theta)$.
\end{theorem}
\begin{proof}
From \cref{lemma:descent_lemma}, we have that
\begin{align*}
    \sum_{i=1}^N p_{t+1, i} \ell_i(\theta_{t+1}) - \sum_{i=1}^N p_{t, i} \ell_i(\theta_{t}) \leq
    -\frac{1}{2\beta} \norm{\sum_{i=1}^N p_{t,i} \nabla \ell_i(\theta_t)}^2.
\end{align*}
Summing over $t$, leads to 
\begin{align*}
    \sum_{i=1}^N p_{T, i} \ell_i(\theta_{T}) - \sum_{i=1}^N p_{0, i} \ell_i(\theta_{0}) \leq
    -\frac{1}{2\beta} \sum_{t=1}^T\norm{\sum_{i=1}^N p_{t,i} \nabla \ell_i(\theta_t)}^2.
\end{align*}
Rearranging the above equation and using the minimality of $p^\star, \theta^\star$, we get that
\begin{align*}
    \frac{1}{T} \sum_{t=1}^T & \norm{\sum_{i=1}^N p_{i,t} \nabla \ell_i(\theta_t)}^2 \leq \\
    & \quad \quad \frac{2\beta }{T}\bigg(\sum_{i=1}^N p_{0,i} \ell_i(\theta_0) - \sum_{i=1}^N p_i^\star \ell_i(\theta^\star)\bigg).
\end{align*}
\end{proof}

\subsection{Proof of \cref{thm:sgd_mw_converge}}
\begin{theorem}
Assume that for $1\leq i \leq N: \ell_i(\cdot)$ is $\beta$-smooth, $G$-Lipshitz $B$-bounded. Define $F(\theta, p) = \E_{i\sim p} \ell_i(\theta)$.
Then when running MR with SGD that samples according to $i_t \sim p_t$ and uses $\alpha=\sqrt{\frac{2B}{G^2\beta T}}$, it holds that:
\begin{align*}
    \frac{1}{T} \sum_{t=1}^T \E_{i_1, ..., i_t} \norm{\nabla F(\theta_t, p_t)}^2 \leq G \sqrt{\frac{2\beta B}{T}}.
\end{align*}
\end{theorem}
\begin{proof}
Note that $F(\theta, p) = \sum_{i=1}^N p_i \ell_i(\theta)$. Using smoothness of $F(\cdot, \cdot)$ w.r.t the first argument and the bounded gradients assumption, we get:
\begin{align*}
    F(\theta_{t+1}, p_{t})
    &\leq F(\theta_t, p_t) 
    -\alpha \nabla F(\theta_t, p_t)(-\nabla \ell_{i_t}(\theta_t)) 
    \\ &+ \frac{\beta}{2} \alpha^2 \norm{\nabla \ell_{i_t}(\theta_t)}^2 \\
    &\leq F(\theta_t, p_t) +\alpha \nabla F(\theta_t, p_t)\nabla \ell_{i_t}(\theta_t) + \frac{\beta\alpha^2 G^2}{2}.
\end{align*}

Notice that SGD with MW can still be seen as an FTRL algorithm with a constant step size, where the losses are fixed from the FTRL point of view. Thus, from \cref{optimality_p}, we have that $F(\theta_{t+1}, p_{t+1}) \leq F(\theta_{t+1}, p_{t})$ holds.

Taking expectation on both sides w.r.t $i_1,..,i_{t} \sim p_1, ..., p_{t}$ yields:
\begin{align*}
    \E_{i_1, ... i_{t}} F(\theta_{t+1}, p_{t+1}) &\leq 
    \E_{i_1, ... i_{t}} F(\theta_t, p_t) 
    \\&+ \E_{i_1, ... i_{t}} \Big[\nabla F(\theta_t, p_t) (-\alpha \nabla \ell_{i_t}(\theta_t))\Big]
    \\&+ \frac{\beta}{2} \alpha^2 G^2.
\end{align*}
We will take a closer look at the expression $\E_{i_1, ..., i_{t}} [\nabla F(\theta_t, p_t) (\nabla \ell_{i_t}(\theta_t))]$. Using the law of total expectation, we have that
\begin{align*}
    \E_{i_1, ..., i_{t}}& [\nabla F(\theta_t, p_t) (\nabla \ell_{i_t}(\theta_t))] =\\
    &= \E_{i_1, ..., i_{t-1}} [\E_{i_t} [\nabla F(\theta_t, p_t) (\nabla \ell_{i_t}(\theta_t))| i_1, ..., i_{t-1}]] \\
    &= \E_{i_1, ..., i_{t-1}} [\nabla F(\theta_t, p_t) \E_{i_t}  [\nabla \ell_{i_t}(\theta_t)| i_1, ..., i_{t-1}]] \\
    &= \E_{i_1, ..., i_{t-1}} [\norm{\nabla F(\theta_t, p_t)}^2] \\
    &= \E_{i_1, ..., i_{t-1},i_t} [\norm{\nabla F(\theta_t, p_t)}^2].
\end{align*}
Plugging this into the earlier expression we get:
\begin{align*}
    \E_{i_1, ... i_{t}} F&(\theta_{t+1}, p_{t+1}) \leq \\
     &\E_{i_1, ... i_{t}} F(\theta_t, p_t) -\alpha\E_{i_1, ... i_{t}} \Big[\norm{\nabla F(\theta_t, p_t)}^2 \Big] \\&+ \frac{\beta}{2} \alpha^2 G^2.
\end{align*}
Summing over $t=1,..,T$ and averaging leads to
\begin{align*}
    \frac{1}{T} \E[&\norm{\nabla F(\theta_{t+1}, p_{t+1})}^2] 
    \leq \\ & \quad \quad \leq \frac{\E F(\theta_T, p_T) - \E F(\theta_1, p_1)}{\alpha T}
    + \frac{\beta}{2} \alpha G^2 \\
    & \quad \quad \leq \frac{B}{\alpha T} + \beta G^2 \alpha.
\end{align*}
By setting $\alpha=\sqrt{\frac{2B}{G^2\beta T}}$, we get the desired bound.
\end{proof}

\subsection{Proof of \cref{eq:logistic_clean_loss}}
\begin{lemma}
When running GD starting with $\theta_0 = 0$, using the clean labels $\bs{y}$  and step size $\alpha>0$, it holds that $\forall \epsilon > 0$ there exists $t > \frac{-\log(\exp(\epsilon) - 1)}{\alpha \exp(-\epsilon) (\exp(\epsilon) - 1)}$ such that
\begin{align*}
    L(\theta_t) = \log(1+\exp(-\theta_t)) < \epsilon.
\end{align*}
\end{lemma}
\begin{proof}
    In the logistic loss in the case we discuss, it hold that $x_i=y_i=1$ for $1\leq i \leq N$. Thus, we have 
    \begin{align}\label{eq:clean_log_loss}
        L(\theta) &= \frac{1}{N} \sum_{i=1}^N \log(1 + \exp(-x_i y_i \theta)) 
        = \log(1+\exp(-\theta)).
    \end{align} 
    The gradient in this case is equal to $\nabla L(\theta_t) = \frac{-1}{1+\exp(\theta_t)}$ and therefore $\theta_{t+1} = \theta_t + \frac{\alpha}{1+\exp(\theta_t)}$. When we set $\theta_0 = 0$, we get a positive and increasing series.
    
    By observing \cref{eq:clean_log_loss}, our goal is to find $\theta_t \in \mathbb{R}, t \geq 1$ s.t $\theta_t \geq -\log(\exp(\epsilon) -1)$.
    Note that for all $t$ s.t $\theta_t < -\log(\exp(\epsilon) -1)$ it holds that
    \begin{align*}
        \theta_{t+1} - \theta_t &= \frac{\alpha}{1+\exp(\theta_t)} 
        \\&\geq \frac{\alpha}{1 + \exp(-\log(\exp(\epsilon) -1))} 
        \\&= \alpha \exp(-\epsilon)(\exp(\epsilon)-1).
    \end{align*}
    Therefore, for $t > \frac{-\log(\exp(\epsilon) -1)}{\alpha \exp(-\epsilon)(\exp(\epsilon)-1}$, it must hold that $L(\theta_t) \leq \epsilon$.
\end{proof}

\subsection{Proof of \cref{lemma:convergence_gd_logistic}}
\begin{lemma}
    When running GD starting with $\theta_0 = 0$, using the noisy labels $\bs{\yT}$ and step size $\alpha=1$, the algorithm converges to $\theta^\star = \log( \frac{1-\sigma}{\sigma})$. Formally,
    for any $\epsilon > 0$ and $T=O(\frac{1}{\epsilon})$ it holds that $\abs{\theta_t-\theta^\star} \leq \epsilon$.
% \end{replemma}
\end{lemma}
\begin{proof} We will first show that     the loss the learner observes is $1$-smooth by bounding its second derivative:
    $\frac{d^2\LT(\theta)}{d \theta} = \frac{\sigma}{(\exp(-\theta)+1)^2} + \frac{1-\sigma}{(\exp(\theta)+1)^2} \leq \sigma + 1 - \sigma = 1$
    Therefore, it holds that $\LT(\theta)$ is $1$-smooth.
    Note that the observed loss is
    \begin{align*}
        \LT(\theta) = \sigma \log(1+\exp(\theta)) + (1-\sigma) \log(1+\exp(-\theta)),
    \end{align*}
    and its gradient is
    \begin{align*}
        \nabla \LT(\theta) = \frac{\sigma}{1+\exp(-\theta)} - \frac{1-\sigma}{1+\exp(\theta)}.
    \end{align*}
    For $\theta^\star = \log(\frac{1-\sigma}{\sigma})$ it holds that $\nabla \LT(\theta^\star) = 0$ and $\nabla^2 \LT(\theta^\star) > 0$ so $\theta^\star$  achieves the minimum. Note that $\theta^\star$ is the only critical point. Since GD converges to a critical point  when running it with $\beta$-smooth function and step size $\alpha=\frac{1}{\beta}$ \cite{bubeck2015convex}, then in this case GD converges to $\theta^\star$.
    
\end{proof}

\subsection{Proof of \cref{lemma:increasing_w_logistic}}
\begin{lemma}
    When running \cref{alg:gd_mw} with the step sizes $\eta=\alpha =1$, $\mu \geq 2$ and $\theta_0 = 0$, then $\forall t \geq 0$  it holds that $\theta_{t+1} > \theta_t$.    
\end{lemma}
\begin{proof}
Note that in each step $t\geq 0$ the losses of all corrupted examples is the same so they share the same probability which we will denote by $p_{cr,t}$. The same holds also for the clean examples' probabilities, which we will denote by $p_{cl,t}$.
Let $\delta_t \coloneqq \theta_{t+1} - \theta_t = \frac{(1-\sigma)p_{cl,t}}{1+\exp(\theta_t)} - \frac{\sigma p_{cr,t}}{1+\exp(-\theta_t)}$. 
Lastly, denote by $A_t = \frac{p_{cl,t+1}}{p_{cl,t}}$ and $B_t = \frac{p_{cr,t+1}}{p_{cr,t}}$ the ratios between two consecutive clean and noisy probabilities.

The proof will hold by induction. For the base step it holds that $\theta_1 = \Delta > 0$.
Next, assume that $\forall s \leq t$: $\delta_s > 0$. Then,
\begin{align*}
    \delta_{t+1} = \frac{(1-\sigma)p_{cl,t}A_t}{1+\exp(\theta_{t+1})} - \frac{\sigma p_{cr,t}B_t}{1+\exp(-\theta_{t+1})}.
\end{align*}
Note that $\frac{B_t}{A_t} = \exp(-\theta_{t+1})$, and therefore
\begin{align} \label{eq:logistig_grad}
    \delta_{t+1} &= A_t\bigg( \frac{(1-\sigma)p_{cl,t}}{1+\exp(\theta_{t+1})} - \frac{\sigma p_{cr,t}\exp(-\theta_{t+1})}{1+\exp(-\theta_{t+1})} \bigg) 
    \\&= \frac{A_t}{1+\exp(\theta_{t+1})}((1-\sigma)p_{cl,t} - \sigma p_{cr,t}).
\end{align}
Since for $s \leq t$ it holds that $\theta_s \geq 0$, we have that $\lT_{cr}(\theta_s) = \log(1+\exp(\theta_s)) \geq \log(1+\exp(-\theta_s)) = \lT_{cl}(\theta_s)$, which yields that $p_{cr,t} \leq p_{cl,t}$. 
In addition, since all the $(1 - \sigma) N \geq \frac{N}{2}$ clean examples share the same weighting it must hold that $p_{cl} \leq \frac{2}{N} \leq \frac{\mu}{N}$. Thus, the upper bound does not limit our analysis.
Therefore, all the terms in the above equation are strictly positive.
\end{proof}

\subsection{Proof of \cref{theorem:gd_mw_logistic}}
\begin{theorem}
    When running \cref{alg:gd_mw} with the step sizes $\eta=\alpha =1$, $\mu \geq 2$ and $\theta_0 = 0$, then for any
     $\epsilon>0$ exists $t \geq \max\Big\{ \frac{-\log(\frac{1}{2})}{\Delta} + 2, \frac{-\log(\exp(\epsilon)-1)}{\exp(-\epsilon)(\exp(\epsilon)-1)} \frac{2N}{\sigma} \Big\}$ such that $L(\theta_t) \leq \epsilon$.
\end{theorem}
\begin{proof} 
    From the MW update rule it holds that $p_{cr,t} = \exp(-\sum_{s=1}^t \theta_s){p_{cl,t}}$. Using $\sigma \leq \frac{1}{2}$ and \cref{eq:logistig_grad}, we have that
    \begin{align*}
        \delta_{t+1} &> \frac{A_t p_{cl,t} \sigma}{1+\exp(\theta_{t+1})}(1-\exp(-\sum_{s=1}^t \theta_s)).
    \end{align*}
    Since $\theta_1 = \Delta > 0$ and by \cref{lemma:increasing_w_logistic} we get that $\theta_t > \Delta$ for $t > 1$.
    In addition, since $\theta_t$ is increasing (\cref{lemma:increasing_w_logistic}), it holds that: $A_t \geq 1$ for $1 \leq t$ so $p_{cl,t} \geq p_{cl,1} = \frac{1}{N}$.
    So overall, we have that
    \begin{align*}
        \delta_{t+1} &\geq \frac{p_{cl,t+1}\sigma}{1+\exp(\theta_{t+1})} (1-\exp(-\Delta(t-1)))
        \\&\geq \frac{\sigma}{N} \frac{1-\exp(-\Delta(t-1))}{1+\exp(\theta_{t+1})}.
    \end{align*}
    For $t \geq \frac{-\log(\frac{1}{2})}{\Delta} + 1$ it must hold that $\delta_{t+1} > \frac{\sigma}{2N (1+\exp(\theta_t))}$. Using the same argument as in the proof of  \cref{eq:logistic_clean_loss} for the unbounded gradients, we get that for $t \geq \max\{ \frac{\log(-\frac{1}{2})}{\Delta} + 2, \frac{-\log(\exp(\epsilon)-1)}{\exp(-\epsilon)(\exp(\epsilon)-1)} \frac{2N}{\sigma} \}$ the loss value satisfies $L(\theta_t) \leq \epsilon$
\end{proof}

\subsection{Proof of \cref{base}}
\begin{lemma}\label{base}
for $i \notin I_{cr}$, we have that 
\begin{align*}
    p_{1, i} \geq (N +N\sigma(\exp(-2\eta\epsilon^2\Delta) - 1)^{-1}.
\end{align*}
\end{lemma}
\begin{proof} 
It holds that $\bs{\yT} = \bs{y} + \bs{\epsilon}$, $\theta^\star = 1$ is the LS solution with $\bs{x\sqrt{P_1}}$ and $\bs{\sqrt{P_1}y}$. The least squares solution to the corrupted problem is
$\theta_1 = \bs{xP_1}(\bs{y} + \bs{\epsilon}) = \bs{xP_1y} + \bs{xP_1\epsilon} = \theta^\star + \bs{xP_1\epsilon}$.

Note that since $L(\theta^\star) = 0$, $\bs{x^T} \theta^\star - \bs{y} = \bs{0}$ the losses satisfy:
\begin{align*}
    i \notin I_{cr}: \quad \lT_i(\theta_1) 
    =\frac{1}{2}(x_i \bs{x P_1\epsilon })^2  
    = \frac{1}{2N^2} (\bs{x \epsilon})^2,
\end{align*}
\begin{align*}
     k \in I_{cr}: \lT_k(\theta_1) 
     &=\frac{1}{2}(x_k \bs{x P_1\epsilon} - \epsilon_k)^2 
     \\&=\frac{1}{2N^2} (\bs{x \epsilon})^2 + \epsilon^2 \mp 2\frac{\epsilon_k}{N} \bs{x \epsilon}.
\end{align*}

We want to show that the losses seen by the learner satisfy $\lT_i(\theta_1) < \lT_k(\theta_1)$. 
This holds when:
\begin{align*}
     \epsilon^2 \geq 2 \abs{\frac{\epsilon_k}{N} \bs{x \epsilon}}.
\end{align*}

For $\sigma < \frac{1}{2}$, we have that
\begin{align*}
     \abs{2 \frac{\epsilon_k}{N} \bs{x\epsilon}}\leq \frac{2\epsilon^2\sigma N}{N} = 2\epsilon^2 \sigma< \epsilon^2
\end{align*}

Overall when $\sigma = \frac{1}{2} - \Delta$: $\lT_k(\theta_1) - \lT_i(\theta_1) \geq \epsilon^2 \Delta$. Then MW will assign lower weights to the corrupted examples.

Denote $A_1 = \frac{1}{2N^2}(\bs{x\epsilon})^2$. Then, we have
\begin{align*}
    \forall i \notin I_{cr}: & \quad w_{1,i} = \exp(-\eta A_1) \\
    \forall k \in I_{cr}: & \quad 0 \leq w_{1,k} \leq \exp(-\eta A_1 -\eta \epsilon^2 \Delta).
\end{align*}

Looking at the non-corrupted examples' probabilities leads to
\begin{align*}
    p_i 
    &= \frac{\exp(-\eta A_1)}{(1-\sigma)N\exp(-\eta A_1) + \sum_{j \in I{cr}} w_{1,j}} \\ 
    &\geq\frac{\exp(-\eta A_1)}{(1-\sigma)N\exp(-\eta A_1) + \sigma N \exp(-\eta A_1 -\eta\epsilon^2\Delta)}\\
    &=\frac{1}{N} \cdot \frac{1}{1+\sigma(\exp(-\eta\epsilon^2\Delta) - 1)} > \frac{1}{N}
\end{align*}
\end{proof}

\subsection{Proof of \cref{lemma:bounded_p}}
\begin{lemma}
For any $t\in [T]$, $\forall i \notin I_{cr}$ it holds that $p_{t,i} \geq ((1-\sigma)N + \sigma N \exp (-\eta \epsilon^2 \Delta t))^{-1}$.    
\end{lemma}
\begin{proof} 
Proof by induction. The induction base is described in \cref{base}. Assume that for $t$ it holds that $p_{t,i} \geq \frac{1}{(1-\sigma)N + \sigma N \exp (-\eta \epsilon^2 \Delta t)}$.
Hence,
\begin{align}\label{sum_clean_p}
    \sum_{i \notin I_{cr}} p_{t,i} \geq \frac{1-\sigma}{1 + \sigma(\exp(-\eta \epsilon^2 \Delta t) -1)} \geq 1 - \sigma.
\end{align} 
Since we have a 1D normalized data with $L(\theta^\star) = 0$, for any probability matrix, $\bs{P}$ the LS solution is $\theta^\star$, and hence we can use that $\theta_t = \theta^\star + \bs{xP_t\epsilon}$. Denote by $A_t = \frac{1}{2}(\sum_{j\in I_{cr}} p_{t,j} x_j \epsilon_j)^2$, then the loss term is:
\begin{align*}
    k \in I_{cr}: &\lT_{k}(\theta_t) = A_t + \frac{1}{2} \epsilon^2 - \epsilon_k \sum_{j\in I_{cr}} p_{t,j} a_j \epsilon_j, \\
    i \notin I_{cr}: & \lT_{i}(\theta_t) = A_t.
\end{align*}
Hence (using \cref{sum_clean_p}),
\begin{align*}
    \forall s\in [t], k\in I_{cr}, i\notin I_{cr}: &\\
    \lT_{k}(\theta_s) - \lT_{i}(\theta_s) &
    \geq \frac{1}{2}\epsilon^2 - \epsilon^2 \sum_{j\in I_{cr}s} p_{s,j} \geq \epsilon^2 \Delta. 
\end{align*}
Finally, since for $j\in I_{cr}$ it holds that  $w_{t+1,j} \leq \exp(-\eta(\sum_{s=1}^{t+1} A_s + \epsilon^2\Delta))$, we have that
\begin{align*}
    p_{t+1,i}
    &=\frac{\exp(-\eta\sum_{s=1}^{t+1} A_s)}{(1-\sigma)N\exp(-\eta\sum_{s=1}^{t+1} A_s) + \sum_{j\in I_{cr}} w_{t+1,j}}
    \\&\geq \frac{1}{(1-\sigma)N + \sigma N \exp(-\eta \epsilon^2 \Delta (t+1))}
\end{align*}
\end{proof}

\subsection{Proof of \cref{theorem:linear}}
\begin{theorem}
 For \cref{alg:ls_mw} we have that for $c >0 $ there exists $t > \frac{\ln(\frac{ \epsilon}{c+1+\Delta})}{\eta \epsilon^2 \Delta}$ such that:
 \begin{align*}
     \abs{\theta_t - 1} \leq c ,
 \end{align*}
 where $\theta_t$ is the learned parameter by the algorithm.    
\end{theorem}
\begin{proof} \label{proof:linear}
We know that for any distribution matrix $\bs{P}$ it holds that $\theta^\star = \bs{xP y }$. Recall that $\theta_t = \bs{xP_t}(\bs{y} + \bs{\epsilon})$. Therefore,
\begin{align*}
    \abs{\theta_t - \theta^\star} 
    &= \abs{\theta^\star  - \bs{x P_t y} - \bs{xP_t \epsilon}} 
    \\&= \abs{\bs{x{P_t}\epsilon}}
    = \abs{\sum_{j=1}^N \epsilon_j p_{t,j}}\leq \epsilon \sum_{j\in I_{cr}} p_{t,j}.
\end{align*}
Using \cref{lemma:bounded_p}, we get that $\sum_{j\in I_{cr} }p_{t,j} \leq \frac{\sigma \exp(-\eta\epsilon^2\Delta t)}{1+\sigma(\exp(-\eta\epsilon^2\Delta t) -1)} \leq \frac{\exp(-\eta\epsilon^2\Delta t)}{1+\Delta}$.
Thus we have that,
\begin{align*}
    \abs{\theta_t - \theta^\star} \leq \epsilon \sum_{j\in I_{cr} }p_{t,j} \leq \frac{\epsilon\exp(-\eta \epsilon^2 \Delta t)}{1+\Delta},
\end{align*}
for $t > \frac{\ln(\frac{\epsilon}{c+1+\Delta})}{\eta \epsilon^2 \Delta}$ therefore it holds that $\abs{\theta^\star - \theta_t} \leq c$
\end{proof}

\subsection{Proof of \cref{lemma:lipschitz}}
\begin{lemma}
For an MW update of $p_{i} = \frac{\exp(-\eta\ell(x_i;y_i,\theta))}{\sum_{j=1}^N \exp(-\eta\ell(x_j;y_j,\theta))}$ (without history) and $y_i\in\{1,-1\}$ it holds that the uniform average of the absolute derivatives is larger than the MR weighting of absolute derivatives,
    \begin{align*}
        \frac{1}{N} \sum_{i=1}^N    \abs{\frac{\partial \ell(x_i;y_i,\theta)}{\partial x_i}} \geq \sum_{i=1}^N p_i \abs{\frac{\partial (x_i;y_i,\theta)}{\partial x_i}}.
        \end{align*}
\end{lemma}

\begin{proof}
    Recall the loss is $\ell(x_i; \theta, y_i) = \log(1+\exp(-x_i \theta y_i))$, shortly we notate it as $\ell_i$. The derivative is 
    $$\frac{\partial \ell_i}{\partial x_i} = \frac{-\theta y_i \exp(-\theta y_i x_i)}{1+\exp(-\theta y_i x_i)} = -\frac{\theta y_i}{\exp(x_i\theta y_i) + 1}.$$
    Next, we establish the connection between the loss and the derivative. 
    For $y_i=1$ it holds $$\ell(x_i;y_i,\theta) = \log(1+\exp(-x_i\theta)) \Rightarrow \exp(-x_i\theta) = \exp(\ell_i) -1 ,$$
    and 
    $$ \frac{\partial \ell_i}{\partial x_i} = -\theta (1-\exp(-\ell_i)).$$
    For $y_i=-1$ it holds
    $$\ell(x_i;y_i,\theta) = \log(1+\exp(x_i\theta)) \Rightarrow \exp(x_i\theta) = \exp(\ell_i) -1 $$
    and $$ \frac{\partial \ell_i}{\partial x_i} = \theta (1-\exp(-\ell_i)).$$
    Overall, the absolute value of the derivative in both cases is $\abs{\theta} (1-\exp(-\ell_i))$ which is monotonly increasing in $\ell_i$.

    Assume w.l.o.g. that $\ell_i$ are sorted, i.e. $\ell_1 \leq \ell_2 \leq ... \leq \ell_N$. Let $k$  be an index such that for all $i\geq k$ it holds $p_i \leq \frac{1}{N}$.
    Note that this is possible due to the monotone weighting update rule $p_i = \frac{\exp(-\eta\ell_i)}{\sum_{j=1}^N \exp(-\eta\ell_j)}$.
    The $k$th to $N$th examples have the largest losses and the lowest weights.

Using the observation that higher loss induces higher absolute derivative value, we have that for $i < k$ is holds that $\abs{\frac{\partial \ell_i}{\partial x_i}}  \leq \abs{\frac{\partial \ell_k}{\partial x_k}}$ and for $i \geq k$ it holds $\abs{\frac{\partial \ell_i}{\partial x_i}} \geq \abs{\frac{\partial \ell_k}{\partial x_k}}$.
Then,
\begin{align*}
    \sum_{i=1}^{k-1} \left(\frac{1}{N} - p_i\right) \abs{\frac{\partial \ell_i}{\partial x_i}} &+ \sum_{i=k}^{N} \left(\frac{1}{N} - p_i\right) \abs{\frac{\partial \ell_i}{\partial x_i}} 
    \\&\geq \sum_{i=1}^{k-1} \left(\frac{1}{N} - p_i\right) \abs{\frac{\partial \ell_k}{\partial x_k}} + \sum_{i=k}^{N} \left(\frac{1}{N} - p_i\right) \abs{\frac{\partial \ell_k}{\partial x_k}} = 0.
\end{align*}

\end{proof}

\section{Additional Experiments and Details} \label{sec:additional_experiments}

Here we provide additional experiments to the ones presented in the paper. Specifically, we show the effect of MR on noisy examples' probabilities, then we show the grid search performed to find the only parameter of our method, the step size, i.e $\eta$ .
Then we experiment MR with Adam and list the results for training in the presence of noisy labels with mixup and label-smoothing. We further show the results for adversarial training with Adam with and without mixup.

{\bf Probabilities and loss with MR.} 
Aiming to analyse the learned weighting we present in \cref{fig:top_low_losses} the loss of the examples with the lowest and highest weights after training with $40\%$ and $20\%$ label noise with CIFAR-10.
It is clear from the results that examples with low weights suffer from high loss while the heaviest examples incurred low loss.

\begin{figure*}
    \centering 
    \begin{subfigure}{0.45\linewidth}
     \includegraphics[width=1\linewidth]{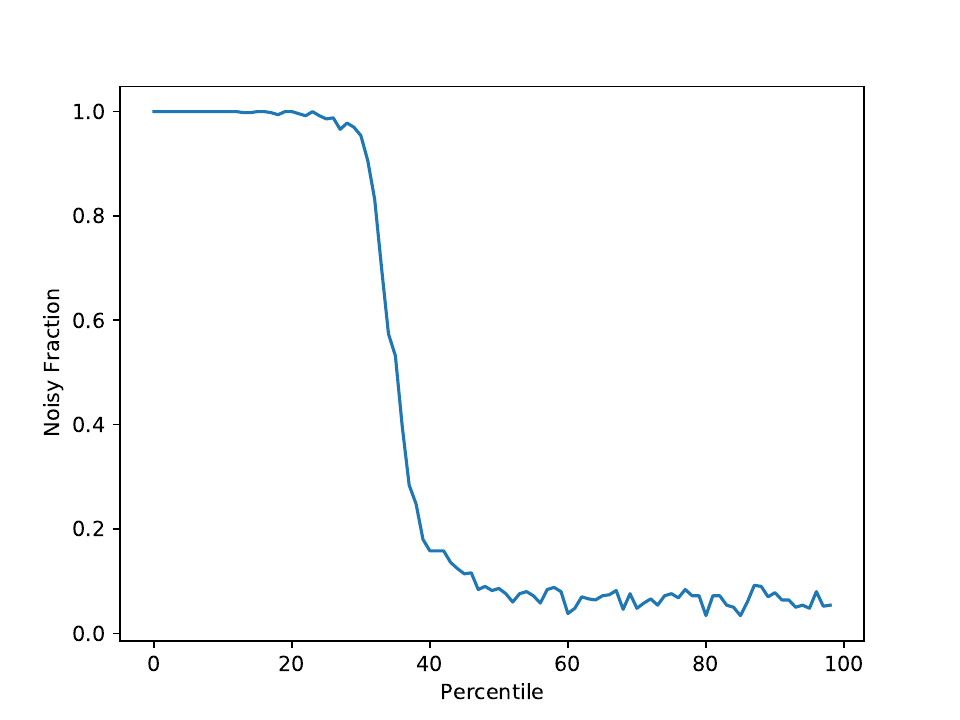}
    \end{subfigure}
    \begin{subfigure}{0.45\linewidth}
     \includegraphics[width=1\linewidth]{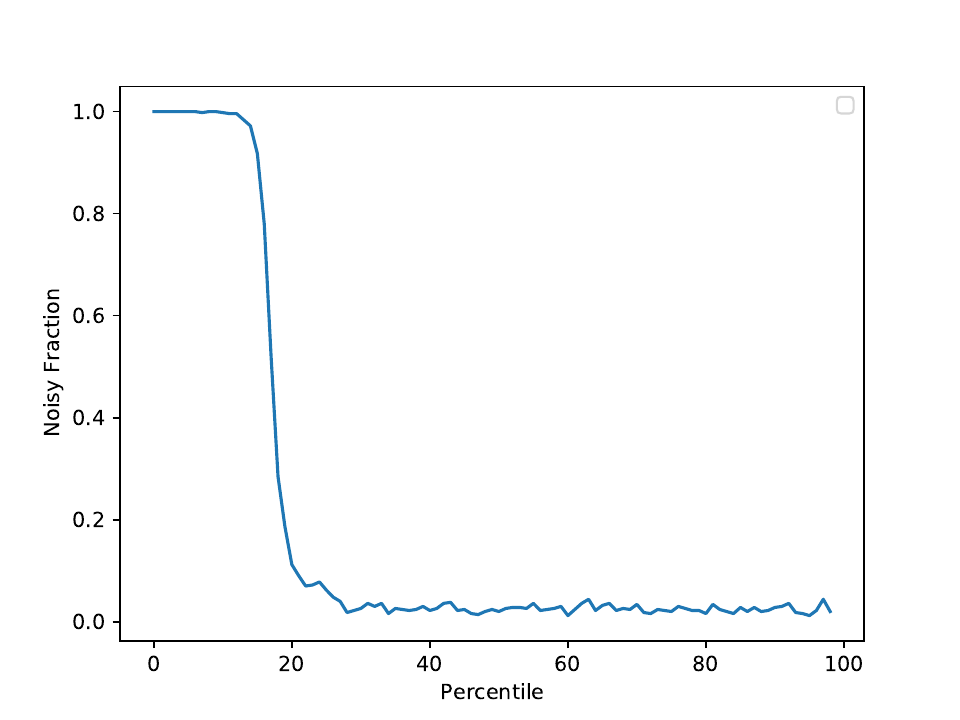}
    \end{subfigure}
  \caption{Fraction of examples with noisy labels along weighting percentiles. Trained with CIFAR10 and (a), (b) are training with 40\% and 20\% noise levels respectively.}
  \label{fig:perc_noisy_frac}
\end{figure*}

In \cref{fig:perc_noisy_frac} we split the training examples to $100$ percentiles according to their weights.
\cref{fig:perc_noisy_frac} shows the fraction of noisy labels per percentile.
It can be seen that indeed noisy labels get lower weights when trained with MR. Almost all the examples in the percentiles which are lower than the noise ratio are noisy.

\begin{figure*}
    \centering
\begin{subfigure}{0.45\linewidth}
    \centering
    \includegraphics[width=1\linewidth]{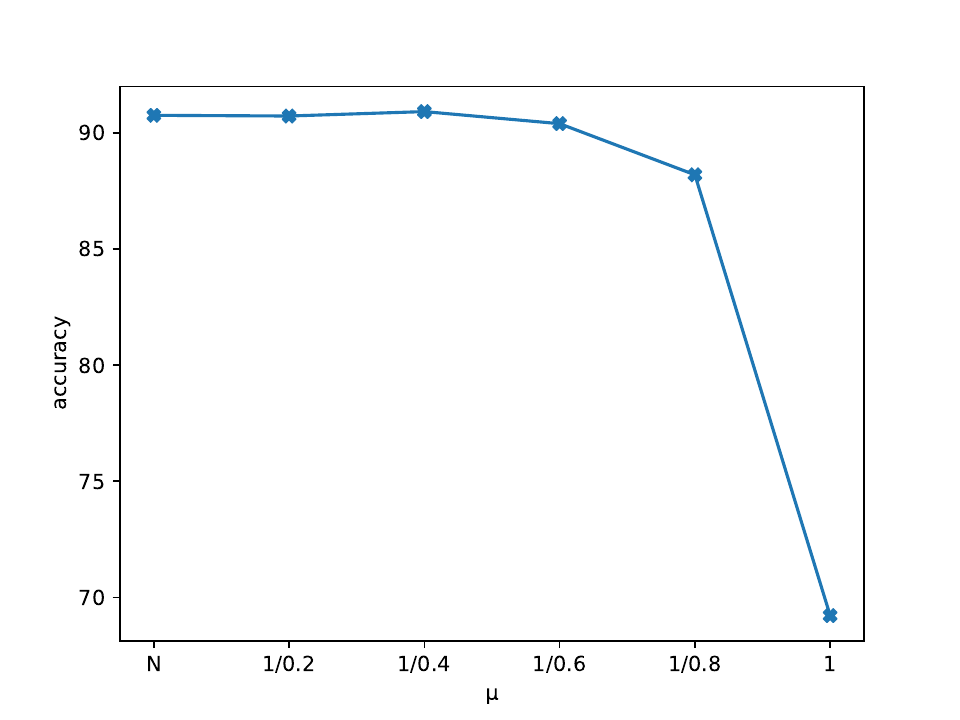}
    \caption{Accuracy results when the weights are bounded.}
    \label{fig:max_weight}
\end{subfigure}
\begin{subfigure}{0.45\linewidth}
    \centering
    \includegraphics[width=1\linewidth]{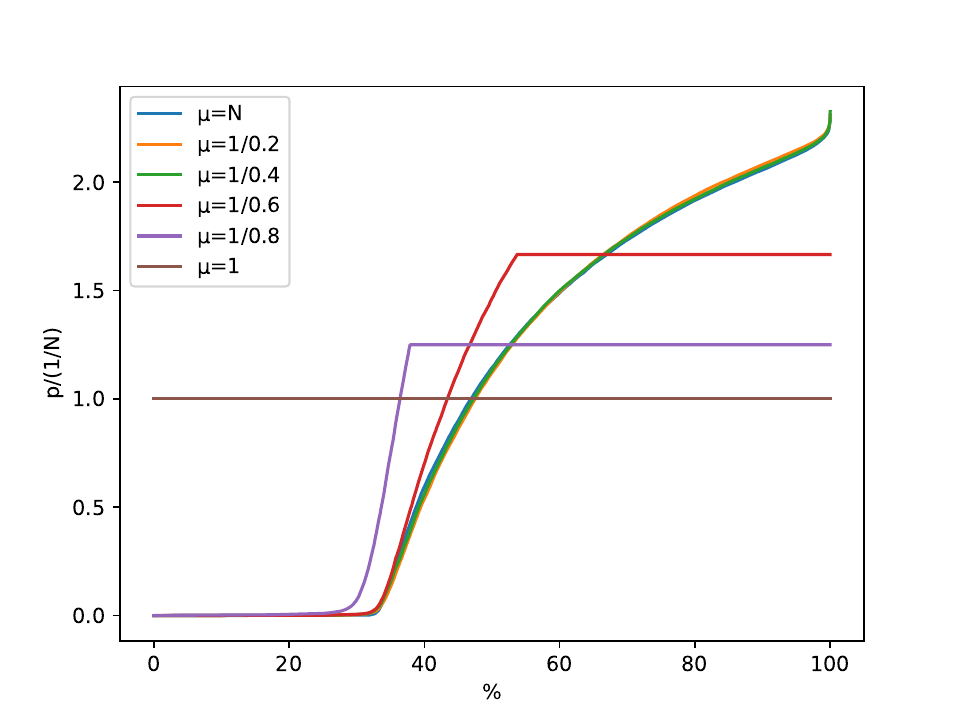}
    \caption{Ratios between learned weights and uniform weights at end of training with multiple $\mu/N$ upper bounds values.}
    \label{fig:ratio_weighting}
\end{subfigure}
\caption{Upper bound effect.} \label{fig:upper_bound}
\end{figure*}

{\bf Effect of upper bound.}
The $\mu/N$ upper bound of the weights is necessary to insure that MR does not converge to degenerated $p$ vectors.
We examine the effect of $\mu$ of the accuracy and the weighting.
In \cref{fig:max_weight} are presented accuracy results of MR with different bounds on the weighting when trained on CIFAR10 with 40\% noisy labels. 
One can see that the accuracy remain quite stable for high $\mu$ values, we see degradation in accuracy when $\mu \geq 1/0.8$ which forces the weighting to have non-zero weights for noisy examples.
In \cref{fig:ratio_weighting} we plot the ratio of the learned weighting with $1/N$. It is shown that $\mu \in \{N, 1/0.2, 1/0.4\}$ the learned distribution is similar. Training with other $\mu$ values change the distribution and one can observe a fraction of the weighting have the maximal allowed value, as $\mu$ is closer to 1 there are more weighting with maximal values.

{\bf Experiments details with sparse regularization \cite{zhou2021learning}. \label{sup:experiment_sr}}
For the experiments with the addition of sparse regularization we use the same setting as in their original paper.
We train 8-layers CNN for CIFAR-10 and ResNet34 for CIFAR-100 for 120 and 200 epochs, respectively.
In addition, we use SGD with momentum 0.9 and cosine learning rate scheduler with 128 batch size.
We employ $\ell_2$ regularization of $10^{-4}$ and $10^-5$ for CIFAR-10 and CIFAR-100 respectively with initial learning rate of $0.01$ and $0.1$. 
For the sparse regularization hyper-parameters we use $(\tau, p, \lambda_0, \rho, r) = (0.5, 0.1, 1.1, 1.03, 1)$ for CIFAR-10 and $(\tau, p, \lambda_0, \rho, r) = (0.5, 0.01, 4, 1.02, 1)$ for CIFAR-100.
For Clothing1M we use the hyper-parameters which are utilized for training with WebVision \cite{li2017webvision} in the original paper, i.e $(\tau, p, \lambda_0, \rho, r) = (0.5, 0.01, 42 1.02, 1)$.

{\bf Experiments details with UNICON \cite{karim2022unicon}}
We use PreAct ResNet18 \cite{he2016deep}, in their setting they employ a warmup stage where they train the network without their method, we employ MR in this stage with $\eta=0.01$.

{\bf Experiments details with S2E \cite{yao2020searching}}
We use the same setting as described in the original paper, with CNN models that are fully detailed in the original paper for CIFAR10 and CIFAR100.
We add MR on top of the of the S2E method and use the weighting of the examples within the batch for the NN's update.
At the end of each epoch we update  the weighting.
We use $\eta = 0.01$ for the MR step size.

{\bf Experiments details with MW-Net \cite{shu2019meta}}
We use WRN28-10 \cite{zagoruyko2016wide}, in this setting we run for 40 epochs. We employ MR for the weighting of the noisy training data with $\eta=0.01$.

\begin{figure}[H]
    \centering 
 \includegraphics[width=0.5\linewidth]{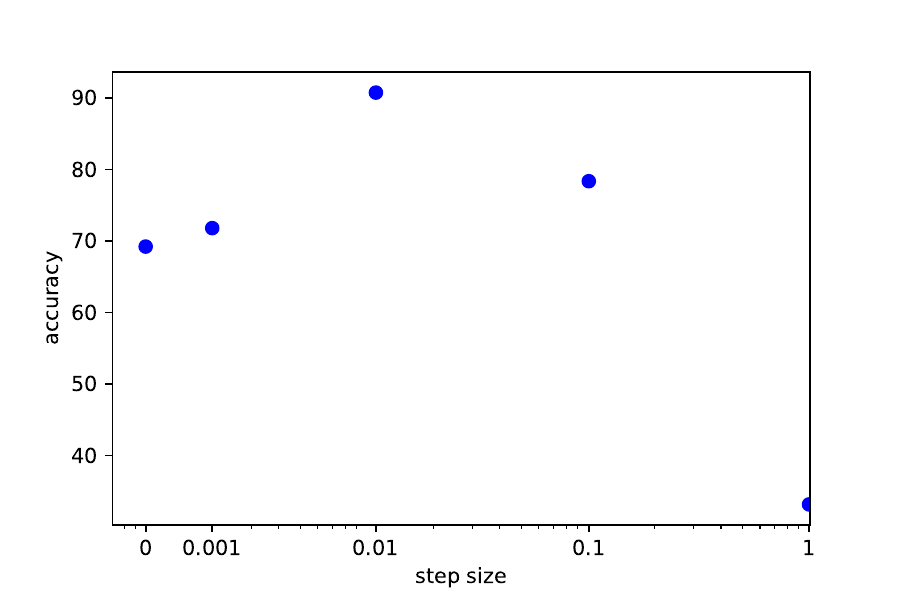}
  \caption{Hyper-parameter grid search of MW step size. }
  \label{fig:grid_search}
\end{figure}

{\bf Hyper-parameter search.}
In \cref{fig:grid_search} we present the hyper-parameter search of the MW step size $\eta$ .
We performed a grid serach over a log scale and tried $4$ different values of $\eta$. We search with CIFAR-10 in the presence of 40\% label noise.
According the results we use $\eta=0.01$ throughout all the experiments mentioned in the paper.

\textbf{Noisy Labels with Adam. } 
We experiment MR when the optimizer, SGD+momentum, is replaced with Adam. Here we use $\alpha=0.5$ for the mixup parameter for improved results.
These results, presented in \cref{tab:label_noise_with_adam}, extend the ones presented in \cref{tab:arch_optim} in the main paper. Here we show also results with mixup, label smoothing, and other noise ratios for CIFAR-10 and CIFAR-100.
Notice that the advantage of using MR increases with the noise level as this method is suitable for improving network robustness in the presence of noise.
When compared to using only Adam with uniform weighting, the addition of MR always improves the results. 
Additionally, using MR with label smoothing improves results in the presence of noise.
Using mixup and the combination of mixup and label smoothing with MR improves the results when the noise ratios are high for CIFAR-10, but for CIFAR-100 we see degradation in accuracy.
\begin{table*}[ht]
\centering
\caption{Results on CIFAR10 and CIFAR100 with different label noise ratio with Adam vs. MR combined with Adam. }
\label{tab:label_noise_with_adam}
\begin{adjustbox}{width=1.\textwidth}
\begin{tabular}{ cc| c c c c c c |c c c c c } 
 \toprule
Dataset                 & Noise ratio   & Adam  & Random Weight. &\shortstack{Adam+\\mixup}                       & \shortstack{Adam+\\label smoothing}                      &\shortstack{Adam+\\mixup+\\label smoothing} & \cite{arazo2019unsupervised}   & MR         & \shortstack{\\ MR + \\mixup}& \shortstack{MR + \\label smoothing} & \shortstack{MR +\\mixup+\\label smoothing} \\
\midrule
\multirow{5}{*}{CIFAR10}& 0\%           & 91.43  & 91.23 & 90.09       & 91.96   & {\bf 92.52}& 88.95 & 91.85       & 91.9        & 91.54       & 91.6             \\
                        & 10\%          & 87.5   & 89.23 & 91.36       & 90.25   & {\bf 91.42}& 85.87 & 91.31       & 89.36       & 91.2        & 91.39                     \\
                        & 20\%          & 84.41  & 87.18 & 90.17       & 86.76   & 90.36      & 85.39 & 90.51       & {\bf 90.89} & 90.46       & 90.8                   \\
                        & 30\%          & 80.55  & 85.43 & 88.29       & 84.21   & 88.34      & 86.63 & 89.67       & {\bf 90.09} & 89.62       & 89.94               \\ 
                        & 40\%          & 72.4   & 83.82 & 86.63       & 80.5    & 86.87      & 82.28 & 88.32       & {\bf 89.36} & 88.11       & 89.11                   \\  
\midrule         
\multirow{5}{*}{CIFAR100}& 0\%         & 69.82   & 68.17 & {\bf 72.88} & 70.43   & 72.37      & 62.18 & 70.54       & 70.89       & 70.97       & 70.66  \\
                        & 10\%         & 59.69   & 62.33 & 67.67       & 62.44   & {\bf 69.26}& 59.88 & 68.84       & 68.34       & 68.86       & 68.55               \\
                        & 20\%         & 52.01   & 58.48 & 64.46       & 54.81   & 64.68      & 58.02 & 66.57       & 67.16       & {\bf 68.13} & 66.07              \\
                        & 30\%         & 44.23   & 53.57 & 61.26       & 47.46   & 61.59      & 51.18 & 64.89       & 63.45       & {\bf 65.19} & 63.45          \\
                        & 40\%         & 37.34   & 44.6  & 52.41       & 39.72   & 59.28      & 49.23 & 61.68       & 60.7        & {\bf 62.66} & 61.61                  \\
\bottomrule
\end{tabular}
\end{adjustbox}
\end{table*}

{\bf Additional comparison with label noise.} We compared MR to two additional methods from \cite{hendrycks2018using,shu2019meta}. We use the same hyper-parameters as mentioned in the original papers and replace the architecture to be ResNet-18 \cite{he2016deep}.
Unlike MR both baselines take advantage of a clean trusted data, we train the model with $1000$ examples that are known to be clean.
In \cref{tab:add_baselines} we see that MR is better in all cases tested even without additional clean validation data.
\begin{table}[H]
    \centering
        \caption{Comparison of MR to additional baselines with CIFAR10. } \label{tab:add_baselines}
    \begin{tabular}{c|ccc}
    \toprule
    Noise Ratio &  \cite{hendrycks2018using} & \cite{shu2019meta} & MR \\ \midrule
    0\%  & 94.28    & 92.24              & {\bf 94.95}           \\
    10\% & 90.65    & 91.34              & {\bf 94.58}       \\
    20\% & 89.51    & 90.9               & {\bf 93.97}  \\
    30\% & 85.71    & 88.5               & {\bf 92.43}           \\
    40\% & 84.9     & 87.37              & {\bf 90.75} \\
    \bottomrule
    \end{tabular}
\end{table}

\textbf{Statistical significance tests.}
We conduct statistical tests to evaluate the significance of the improvement resulting from the addition of MR.
We compare the performance of three baseline models: CE (Base), label smoothing, and mixup, to the same models when MR is included.
To determine statistical significance, we calculate p-values using a two-independent-samples t-test based on three identical and independent experiments.
The results are presented in \cref{tab:p_values} and in most the cases tested the improvement is significant with $p < 0.05$.

\begin{table*}[t]
\centering
\caption{The results of significant tests (p-value) on CIFAR-10/100 with different noise levels.} \label{tab:p_values}

\begin{tabular}{ cc| c c c c} 
 \toprule
Data                 & \shortstack{Noise \\ ratio}   & Base & Smoothing & Mixup   & \shortstack{Smoothing\\+ Mixup}                  \\
\midrule
\multirow{2}{*}{\small CIFAR10}& 20\%    & 0.0000 & 0.0000 & 0.0002  & 0.0000 \\
                        & 40\%   & 0.0000 & 0.0000 & 0.0000  & 0.0000     \\ 
\midrule
\multirow{2}{*}{\small CIFAR100}& 20\%   & 0.0096 & 0.0228 & 0.1998 & 0.3467\\
                         & 40\%  & 0.0001 & 0.0159 & 0.0086 & 0.0100\\
\bottomrule
\end{tabular}

\end{table*}

\textbf{Hyper-parameters for Adversarial Robustness.}
For the adversarial robustness experiments we use ResNet-34 \cite{he2016deep}.
For Free Adversarial Training \cite{shafahi2019adversarial} we train with $m=4$ updates in each batch with CIFAR-10 and with $m=6$ for CIFAR-100.
We train the network for $200$ epochs with $l_2$ regularization of $10^{-3}$, momentum of $0.9$, batch size of $256$ and initial learning rate of $0.1$ which is reduced on plateaus with a factor $0.9$ and patience $3$. We use MR step size of $0.01$.
For mixup the same hyper-parameters as in artificial label noise are employed.

We use an additional baseline, TRADES \cite{zhang2019theoretically} with the same hyper-parameters appear in the original paper: $10$ internal optimization steps, $0.007$ internal step-size and $1/\lambda= 6$.
We train the network for $100$ epochs for CIFAR-10 and $140$ for CIFAR-100, weight decay of $2\times 10^{-4}$, batch size of $128$ and MR step $0.001$.
We apply an initial learning rate of $0.1$ which is multiplied by $0.1$ at epochs $75, 90$ and $100$.

We use Adversarial Robustness Toolbox \cite{nicolae2018adversarial} for the attacks. The PGD attack with $\epsilon \in \{0.01, 0.02, 8/255\}$ is produced with step sizes of $0.003, 0.005, 2/255$, respectively, and maximal iteration of $100$. For FGSM attack we employ step size of $0.001$. The same attack batch size of $256$ is applied.

\textbf{Adversarial Adversarial Robustness with MR.}
Following the listing of the setting's details, we include here the results of MR with adversarial training methods, as mentioned in the main paper.

\textbf{Adversarial Robustness with Adam.}
\cref{tab:adv_adam} shows the results when the original optimizer SGD + momentum is replaced with Adam for CIFAR-10 and CIFAR-100.
We use $\alpha=0.1$ for the mixup parameter, $\eta=0.001$ MR step size and on each mini batch we perform $m=4$ adversarial training steps since we found that those suits better for training with Adam.
It can been seen that adding MR to adversarial training improves results also when the training is performed with Adam.
With the change of optimizer, adding mixup to adversarial training degrades the results and adding MR to this combination does not mitigate the degradation.
The results with Adam reflect the high sensitivity of adaptive methods against attacks with lower accuracy results  compared to training with momentum.
\begin{table*}[ht]
\centering
\caption{Adversarial robustness to attacks with CIFAR-10 with Adam vs. MR combined with Adam; Adv. stands for robust training \cite{shafahi2019adversarial}.} \label{tab:adv_adam}
\begin{adjustbox}{width=1.\textwidth}
\begin{tabular}{ lc c cc c c c| ccccccc} 
\toprule
                & \multicolumn{7}{c|}{CIFAR10}                                             & \multicolumn{7}{c}{CIFAR100}                                                 \\\midrule
Method          & Natural Images& PGD           & PGD         &  PGD       & FGSM      & FGSM        & FGSM       & Natural Images& PGD           & PGD      &  PGD         & FGSM          & FGSM        & FGSM    \\
$\epsilon$      & $0$           & $0.01$        & $0.02$      &  $8/255$   & $0.01$    & $0.02$      & $8/255$    & $0$           & $0.01$        & $0.02$   &  $8/255$     & $0.01$        & $0.02$      & $8/255$ \\\midrule
Adam            & 91.84         & 11.86         & 5.53        & 5.21       & 37.9      & 21.88       & 12.59      & 68.81         & 14.26         & 11.26    & 10.75        & 25.86         & 20.99       & 16.31        \\
Adam+MR         & 91.52         & 13.87         & 5.39        & 5.73       & 42.98     & 24.57       & 15.49      & 69.23         & 13.67         & 10.32    & 9.85         & 26.78         & 20.17       & 14.74        \\
Adam+mixup      & {\bf 92.67}   & 12.32         & 5.01        & 4.65       & 55.33     & 45.84       & 41.68      & {\bf 72.8}    & 15.85         & 10.4     & 9.03         & 35.86         & 30.23       & 27.69        \\ 
Adam+mixup + MR & 90.95         & 11.07         & 5.92        & 5.31       & 51.3      & 38.65       & 33.01      & 65.91         & 10.49         & 8.65     & 8.46         & 28.42         & 20.82       & 17.85        \\\midrule
Adam+Adv.       & 84.19         & 74.92         & 55.31       & 32.47      & 76.61     & 62.98       & 47.91      & 59.12         & 48.28         & 32.71    & 19.21        & 50.72         & 38.48       & 28.72         \\
Adam+Adv.+MR    & 85.1          & {\bf 75.9}    & 56.56       & {\bf 34.26}&{\bf 77.15}& {\bf 63.36} & {\bf 49.2} & 58.36 	      & 48.89	      & 32.95    &  19.49 	    & 50.24	        & 38.52       & 28.51             \\
Adam+Adv.+mixup & 80.15         & 73.45         & {\bf 57.07} & 33.86      & 74.17     & 62.62       & 49.17      & 57.02	      & 47.88	      & 32.98	 & 19.63        & 49.02	        & 38.5        & 28.52        \\
Adam+Adv.+MR+mixup& 84.62       & 75.05         & 56.9        & 31.19      & 76.36     & 61.88       & 46.38      &57.25	      & {\bf 49.51}   & {\bf 35.16}& {\bf 21.79}& {\bf 50.83}   & {\bf 39.44} & {\bf 30.7}        \\
\bottomrule     
\end{tabular}
\end{adjustbox}
\end{table*}

\bibliographystyle{siamplain}
\bibliography{mr.bib}

\end{document}